\pdfoutput=1
\documentclass[11pt,a4paper,twocolumn]{paper}
\usepackage{fix-cm}
\usepackage[all]{hypcap}
\usepackage[authoryear,round]{natbib}

\definecolor{title_blue}{HTML}{204899}
\definecolor{cite_blue}{HTML}{044dc1}
\definecolor{cite_purple}{HTML}{7406a7}
\hypersetup{
  colorlinks=true,
  citecolor=cite_blue,
  linkcolor=cite_purple,
  urlcolor=cite_purple,
  pdftitle={Rationale Distribution Distillation via Flow Matching for Distributional Continuous Latent Reasoning},
  pdfauthor={Mingyeong Kim; Gyusang Cho; Juho Lee; Giung Nam}
}

\usepackage{bm}

\makeatletter
\let\unshifted@maketag\maketag@@@
\renewcommand{\maketag@@@}[1]{%
  \hbox{\kern-3pt\unshifted@maketag{#1}\kern3pt}%
}
\makeatother

\usepackage{array}
\newlength{\analysislabelwidth}
\newlength{\mpscolumnwidth}
\usepackage{tikz}
\usepackage{tikzpagenodes}
\usepackage{placeins}
\usetikzlibrary{arrows.meta,calc,tikzmark}

\newcommand{\lossframe}[2]{%
  \tikzmarknode[outer sep=0pt,minimum height=0pt]{loss-#1}{#2}%
  \tikz[remember picture,overlay]{%
    \path let \p1=(loss-#1.center),\p2=(current page text area.center) in
      \pgfextra{%
        \ifdim\x1>\x2
          \gdef\losscolumn{2}%
        \else
          \gdef\losscolumn{1}%
        \fi
      };
    \path[fill=black,fill opacity=0.05,draw=black!35,
      line width=0.4pt,rounded corners=2pt]
      ([xshift=0.2pt,yshift=4pt]current page column \losscolumn\space area.west |- loss-#1.north)
      rectangle ([xshift=-0.2pt,yshift=-4pt]current page column \losscolumn\space area.east |- loss-#1.south);
  }%
}

\newsavebox{\CBox}
\newcommand{\BF}[1]{%
  \sbox{\CBox}{#1}%
  \resizebox{\wd\CBox}{\ht\CBox}{\textbf{#1}}%
}
\newcommand{\UL}[1]{\underline{#1}}
\newcommand{\BL}[1]{\BF{\UL{#1}}}

\title{Rationale Distribution Distillation via Flow Matching\\for Distributional Continuous Latent Reasoning}

\author[1]{Mingyeong Kim}
\author[1]{Gyusang Cho}
\author[1]{Juho Lee}
\author[1,$\dagger$]{Giung Nam}
\affil[1]{Korea Advanced Institute of Science and Technology}
\correspondingauthor{Correspondence to: \email{giung@kaist.ac.kr}.}

\begin{abstract}
Sampling multiple chain-of-thought rationales from a language model can uncover correct solutions missed by a single attempt.
However, these rationales are typically lengthy, making repeated autoregressive generation costly.
Continuous latent reasoning offers a compact alternative, but point-valued methods condition repeated answer sampling on a single rationale, which can limit exploration of diverse paths and answer coverage.
We thus propose rationale distribution distillation (RDD), training a conditional flow over fixed-length latent sequences by distilling from multiple textual reasoning paths.
Furthermore, we expand the distillation targets through within-query mixup, interpolating between complete latent rationales for the same query.
The flow nonautoregressively generates the entire latent sequence, and a frozen language model autoregressively decodes answers from sampled rationale contexts.
Empirically, the resulting distributional reasoner achieves higher answer coverage than the evaluated point-valued latent reasoning baselines, particularly on advanced mathematical tasks.
\end{abstract}

\begin{document}
\maketitle
\section{Introduction}
\label{sec:introduction}

Recent advances in language models have shown that generating an intermediate rationale, rather than producing an answer directly from a query, can substantially improve performance~\citep{wei2022chain}.
This \emph{chain-of-thought (CoT)} paradigm has since become a standard component of state-of-the-art reasoning models~\citep{openai2024learning,guo2025deepseek-r1}.
However, generating such rationales introduces a significant inference bottleneck.
In particular, CoT traces are often excessively verbose and lengthy~\citep{chen2025do}, leading to substantial overhead in both inference latency and memory consumption.
Such costs have motivated a growing body of work on shortening CoT while preserving reasoning accuracy, with the goal of improving inference efficiency~\citep{han2025token,ma2025cot,munkhbat2025self,kang2025c3ot}.

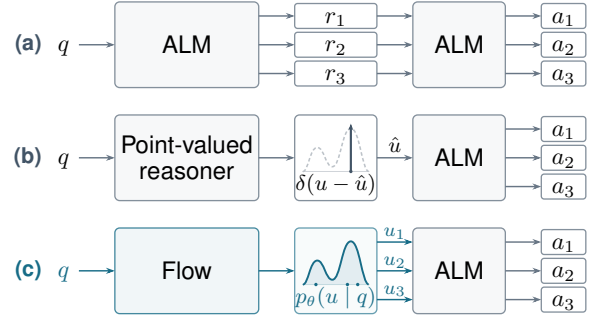
\begin{figure}[t]
  \centering
  \begingroup
  \definecolor{reasoningInk}{HTML}{425466}
  \definecolor{reasoningFlow}{HTML}{176C87}
  \resizebox{\linewidth}{!}{%
    \begin{tikzpicture}[
      x=1cm, y=1cm,
      line/.style={draw=reasoningInk!80, line width=.6pt},
      arrow/.style={line, -{Stealth[length=1.3mm,width=.9mm]}},
      latent arrow/.style={arrow, draw=reasoningFlow},
      panel/.style={
        draw=reasoningInk!65, fill=white,
        rounded corners=2pt, line width=.55pt,
        minimum width=1.2cm, minimum height=1.25cm,
        inner sep=0pt, outer sep=0pt
      },
      model/.style={
        panel, minimum width=1.35cm, fill=reasoningInk!4,
        font=\sffamily\fontsize{10}{11}\selectfont,
        text=black, align=center
      },
      generator/.style={model, minimum width=2.1cm},
      sample/.style={
        draw=reasoningInk!65, fill=white,
        rounded corners=1.2pt, line width=.55pt,
        minimum width=1.2cm, minimum height=.31cm,
        font=\fontsize{10}{11}\selectfont,
        text height=1.6ex, text depth=.5ex,
        inner sep=0pt, outer sep=0pt
      },
      answer sample/.style={sample, minimum width=.65cm},
      query/.style={sample, minimum width=.45cm, draw=none, fill=none},
      heading/.style={
        font=\sffamily\fontsize{8.5}{9.5}\selectfont,
        text=reasoningInk, align=center,
        inner sep=0pt, outer sep=0pt
      },
      panel label/.style={
        heading, anchor=east,
        font=\sffamily\bfseries\fontsize{9}{10}\selectfont
      },
      density/.style={
        insert path={
          (-.48,-.20)
          .. controls (-.40,-.20) and (-.38,.15) .. (-.27,.15)
          .. controls (-.17,.15) and (-.15,-.13) .. (-.05,-.13)
          .. controls (.07,-.13) and (.09,.43) .. (.22,.43)
          .. controls (.35,.43) and (.38,-.18) .. (.48,-.20)
        }
      }
    ]
      \path[use as bounding box] (-.60,0) rectangle (7.90,4.70);

      \def\textRow{4.00}
      \def\pointRow{2.35}
      \def\latentRow{.70}
      \def\queryX{.20}
      \def\arrowGap{.525}
      \pgfmathsetmacro{\generatorX}{\queryX + .45/2 + \arrowGap + 2.1/2}
      \pgfmathsetmacro{\rationaleX}{\generatorX + 2.1/2 + \arrowGap + 1.2/2}
      \pgfmathsetmacro{\decoderX}{\rationaleX + 1.2/2 + \arrowGap + 1.35/2}
      \pgfmathsetmacro{\answerX}{\decoderX + 1.35/2 + \arrowGap + .65/2}
      \node[panel label] at (-.12,\textRow) {(a)};
      \node[panel label] at (-.12,\pointRow) {(b)};
      \node[panel label, text=reasoningFlow] at (-.12,\latentRow) {(c)};

      \node[query] (textQuery) at (\queryX,\textRow) {$q$};
      \node[generator] (teacher) at (\generatorX,\textRow) {ALM};
      \node[model] (textDecoder) at (\decoderX,\textRow) {ALM};

      \node[query] (pointQuery) at (\queryX,\pointRow) {$q$};
      \node[generator] (pointReasoner) at (\generatorX,\pointRow)
        {Point-valued\\reasoner};
      \node[model] (pointDecoder) at (\decoderX,\pointRow) {ALM};
      \node[panel] (pointState) at (\rationaleX,\pointRow) {};

      \node[query, text=reasoningFlow] (latentQuery) at (\queryX,\latentRow) {$q$};
      \node[generator, draw=reasoningFlow!80, fill=reasoningFlow!6]
        (flow) at (\generatorX,\latentRow) {Flow};
      \node[model] (latentDecoder) at (\decoderX,\latentRow) {ALM};
      \node[panel, draw=reasoningFlow!65, fill=reasoningFlow!3]
        (distribution) at (\rationaleX,\latentRow) {};
      \draw[arrow] (textQuery.east) -- (teacher.west);
      \draw[arrow] (pointQuery.east) -- (pointReasoner.west);
      \draw[arrow] (pointReasoner.east) -- (pointState.west);
      \draw[latent arrow] (latentQuery.east) -- (flow.west);
      \draw[latent arrow] (flow.east) -- (distribution.west);

      \draw[arrow] (pointState.east) -- (pointDecoder.west)
        node[midway, above=2pt, font=\fontsize{9}{10}\selectfont,
          inner sep=0pt] {$\hat u$};
      \foreach \offset/\index in {.42/1,0/2,-.42/3} {
        \node[sample] (r-\index) at (\rationaleX,{\textRow+\offset}) {$r_{\index}$};
        \node[answer sample] (text-a-\index) at (\answerX,{\textRow+\offset}) {$a_{\index}$};
        \node[answer sample] (point-a-\index) at (\answerX,{\pointRow+\offset}) {$a_{\index}$};
        \node[answer sample] (latent-a-\index) at (\answerX,{\latentRow+\offset}) {$a_{\index}$};
        \draw[arrow] ([yshift=\offset cm]teacher.east)
          -- (r-\index.west);
        \draw[arrow] (r-\index.east)
          -- ([yshift=\offset cm]textDecoder.west);
        \draw[arrow] ([yshift=\offset cm]textDecoder.east)
          -- (text-a-\index.west);
        \draw[arrow] ([yshift=\offset cm]pointDecoder.east)
          -- (point-a-\index.west);
        \draw[latent arrow] ([yshift=\offset cm]distribution.east)
          -- ([yshift=\offset cm]latentDecoder.west)
          node[midway, above=1pt, text=reasoningFlow,
            font=\fontsize{8}{9}\selectfont, inner sep=0pt]
            {$u_{\index}$};
        \draw[arrow] ([yshift=\offset cm]latentDecoder.east)
          -- (latent-a-\index.west);
      }

      \begin{scope}[shift={(pointState.center)}]
        \draw[draw=reasoningInk!25, line width=.4pt]
          (-.49,-.20) -- (.49,-.20);
        \draw[density, draw=reasoningInk!35, line width=.65pt,
          dash pattern=on 1.5pt off 1pt];
        \draw[-{Stealth[length=1mm,width=.7mm]},
          draw=reasoningInk, line width=.9pt] (.22,-.20) -- (.22,.50);
        \fill[reasoningInk] (.22,-.20) circle[radius=.7pt];
        \node[font=\fontsize{9}{10}\selectfont, inner sep=0pt]
          at (0,-.39) {$\delta(u-\hat u)$};
      \end{scope}

      \begin{scope}[shift={(distribution.center)}]
        \draw[draw=reasoningFlow!40, line width=.4pt]
          (-.49,-.20) -- (.49,-.20);
        \path[density, fill=reasoningFlow!12] -- cycle;
        \draw[density, draw=reasoningFlow, line width=.8pt];
        \foreach \position in {-.29,.17,.32}
          \fill[reasoningFlow] (\position,-.20) circle[radius=.7pt];
        \node[text=reasoningFlow, font=\fontsize{9}{10}\selectfont,
          inner sep=0pt] at (0,-.39) {$p_{\theta}(u\mid q)$};
      \end{scope}
    \end{tikzpicture}%
  }
  \endgroup
  \caption{Comparison of reasoning representations.
    (a) Textual reasoning samples multiple rationales and generates an answer
    for each. (b) Point-valued reasoning samples multiple answers conditioned on one answer context.
    (c) Our flow induces a distribution over answer contexts, enabling diverse answer generation.}
  \label{fig:continuous-reasoning}
\end{figure}

While encouraging language models to generate shorter textual rationales is an intuitive remedy, the \emph{continuous latent reasoning} paradigm offers a more fundamental alternative by reconsidering how intermediate reasoning is represented.
Rather than requiring every reasoning step to be verbalized as a sequence of discrete language tokens, these approaches carry intermediate reasoning through continuous latent states without committing each step to a single discrete token, potentially enabling shorter and more compact reasoning trajectories~\citep{hao2025training,zhang2025soft}.
Such continuous representations may preserve a \emph{superposition} of multiple candidate reasoning states, allowing each latent step to jointly capture information associated with several alternative reasoning paths~\citep{zhu2025reasoning,gozeten2026continuous}.

Prior work supports latent superposition, but the evidence largely concerns synthetic settings~\citep{zhu2025reasoning,gozeten2026continuous}.
Extending this promise to complex tasks is not straightforward, since existing latent methods can struggle to exploit multiple reasoning paths~\citep{suleymanzade2026limits,wu2026llms}.
This limitation matters because exploring alternative paths can uncover correct answers missed by individual attempts~\citep{chen2021evaluating,cobbe2021training}.
We often assess this ability through \emph{answer coverage}, measured by pass@$k$, the probability of finding a correct answer across $k$ independent rollouts.
To learn these alternatives explicitly, we propose \emph{Rationale Distribution Distillation (RDD)}, distilling multiple answer-correct textual CoTs into a query-conditioned latent distribution rather than one fixed latent rationale.

\begin{itemize}[noitemsep,topsep=4pt plus1pt minus2pt,partopsep=0pt]
\item We introduce Rationale Distribution Distillation (RDD), transferring a query-conditioned distribution of textual reasoning paths into latent space. RDD samples a complete latent rationale for each answer rollout, enabling exploration of alternative contexts rather than reusing one fixed latent state (cf. Figure~\ref{fig:continuous-reasoning}).
\item We construct distillation targets from multiple answer-correct CoTs per query. Each trace is compressed separately into a fixed-length latent sequence, then within-query mixup interpolates between complete sequences with a shared weight, augmenting the reference distribution while retaining the original targets.
\item We distill the reference distribution into a conditional flow using flow matching. Independent noise draws yield rationale samples, and each flow step updates all latent positions jointly. This replaces token-by-token rationale generation with parallel refinement while keeping answer decoding autoregressive.
\item We make distributional distillation answer-aware with local content and answer supervision. A contextualizer maps sampled rationales to contexts for a frozen language model that decodes textual answers autoregressively. Answer gradients through this decoder jointly train the contextualizer and flow sampler.
\end{itemize}

\section{Related Work}\label{sec:related-work}

Here, we focus on research directly related to latent reasoning in modern language models.
See Appendix~\ref{sec:additional-related-work} for a broader review of related work.

\paragraph{Point-valued latent reasoning}
Early work explored reasoning beyond explicit text through pause tokens~\citep{goyal2024think}, looped transformers~\citep{giannou2023looped}, and implicit CoT distillation into hidden layers~\citep{deng2023implicit}.
Building on these, Coconut~\citep{hao2025training} established a continuous-input formulation of autoregressive reasoning, feeding model output representations back as next-step inputs in place of token embeddings.
Subsequent approaches augment this formulation with hidden-state self-distillation in CODI~\citep{shen2025codi}, step-level supervision through an auxiliary decoder in SIM-CoT~\citep{wei2026sim}, and span-level distillation using position-weighted token distributions in MUX~\citep{suleymanzade2026mux}.
Furthermore, to reduce sequential computation, PCCoT~\citep{wu2025parallel} iteratively updates latent tokens in parallel using Jacobi iteration, and LOTUS~\citep{fan2026bridging} refines latent blocks with a looped transformer under parallel supervision on gold CoT tokens.
These methods produce a single continuous rationale per query, corresponding to the point-valued reasoner in Figure~\ref{fig:continuous-reasoning}(b).

\paragraph{Distributional latent reasoning}
A separate line of work models or induces distributions over continuous rationales given a query.
Coconut-style reasoners can become stochastic through Monte Carlo dropout or Gaussian noise~\citep{you2026parallel}, while GTS~\citep{wang2026gts} learns context-dependent Gaussian noise with a frozen backbone.
Soft Thinking~\citep{zhang2025soft} deterministically mixes token embeddings using next-token probabilities.
Stochastic Soft Thinking~\citep{wu2026llms} randomizes these weights with Gumbel-Softmax, while Multiplex Thinking~\citep{tang2026multiplex} samples tokens and aggregates their embeddings at each step.
CoLaR~\citep{tan2025think} predicts distributions over embeddings of compressed textual CoT spans.
LaDiR~\citep{kang2026ladir} combines variational autoencoding with flow matching to learn latent CoT block distributions.
Unlike these methods, we explicitly distill each query's empirical distribution of complete textual CoT rationales into a conditional distribution over continuous rationales.
We generate full rationales non-autoregressively and freeze the answer model, as in Figure~\ref{fig:continuous-reasoning}(c).

\section{Method}
\label{sec:method}

\subsection{Textual CoT and answer coverage}
\label{sec:textual-cot-and-answer-coverage}

We first formalize textual CoT sampling, which supplies the sequence-level targets for our continuous rationale generator.
For a \emph{query} $q$, an autoregressive language model (ALM) samples a complete textual \emph{rationale} $r$, then an \emph{answer} $a$ conditioned on both $q$ and $r$~\citep{wei2022chain,kojima2022large}.
The rationale provides an explicit context of intermediate results for answer prediction~\citep{ling2017program,nye2021show}.
Each rollout follows $p(r,a\mid q)=p(r\mid q)p(a\mid r,q)$, inducing an answer distribution that marginalizes over the space $\mathcal R$ of complete textual rationales:
\begin{equation}\textstyle
p_{\mathrm{CoT}}(a\mid q)
=\sum_{r\in\mathcal R}p(a\mid r,q)\,p(r\mid q).
\label{eq:cot-marginal}
\end{equation}
Independent rationale--answer rollouts draw candidate answers from this marginal, with each rollout sampling its own rationale before generating an answer.
We assess the resulting candidate set using pass@$k$, the probability that at least one of $k$ rollouts produces a correct extracted answer~\citep{chen2021evaluating,cobbe2021training}.
This connects answer coverage to sampling complete reasoning paths, rather than repeatedly decoding answers from a single fixed rationale.
In the textual process, however, each typically lengthy rationale must be generated token by token.
Our continuous counterpart retains the rationale--answer sampling structure but updates all rationale positions jointly, reducing sequential dependence during rationale generation while leaving answer decoding autoregressive.

\subsection{Continuous CoT and superposition}
\label{sec:continuous-cot-and-superposition}

Soft Thinking~\citep{zhang2025soft} seeks to reduce the cost of multi-path reasoning by carrying alternative paths in a single continuous trajectory.
At each step, it feeds a probability-weighted mixture of token embeddings into the model, aiming to retain alternatives without separate textual rollouts.
However, recent analysis of Soft Thinking finds that subsequent predictions are dominated by the highest-probability token, making the trajectory resemble greedy decoding~\citep{wu2026llms}.
This suggests that token-level mixtures do not necessarily preserve the diversity of complete rationales that independently sampled textual rollouts can explore.

Coconut~\citep{hao2025training} replaces intermediate text with continuous hidden states fed back as reasoning inputs.
This provides a compact medium in which alternative paths could coexist without being verbalized separately.
Such a representation could support learning from multiple valid paths for the same query.
However, recent controlled multi-chain experiments find that Coconut and CODI do not benefit from additional valid textual CoTs in the evaluated settings~\citep{suleymanzade2026limits}.
These results suggest that access to multiple valid traces alone does not guarantee gains for point-valued reasoners.
Section~\ref{sec:additional-results} revisits this question under our answer-coverage evaluation.

\subsection{Rationale distribution distillation}
\label{sec:rationale-distribution-distillation}

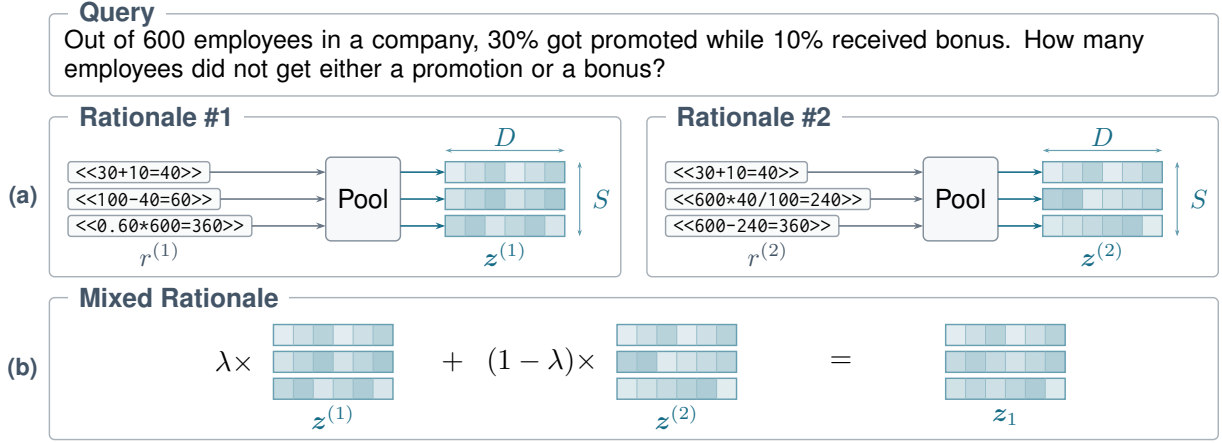
\begin{figure*}[t]
  \centering
  \begingroup
  \definecolor{targetInk}{HTML}{425466}
  \definecolor{targetFlow}{HTML}{176C87}
  \newcommand{\targetRows}[1]{%
    \foreach \row in {0,1,2} {
      \pgfmathsetmacro{\cy}{.36-.36*\row}
      \foreach \col in {0,...,5} {
        \pgfmathtruncatemacro{\shadeA}{12+mod(9*\col+7*\row,27)}
        \pgfmathtruncatemacro{\shadeB}{\row==0 ? \shadeA : 12+mod(5*\col+11*\row+8,27)}
        \pgfmathtruncatemacro{\shade}{#1==1 ? \shadeA : (#1==2 ? \shadeB : (\shadeA+\shadeB)/2)}
        \fill[targetFlow!\shade]
          ({.2666667*\col},{\cy-.14}) rectangle ++(.2666667,.28);
      }
      \draw[draw=targetFlow!65, line width=.55pt]
        (0,{\cy-.14}) rectangle (1.60,{\cy+.14});
      \foreach \col in {1,...,5}
        \draw[draw=targetFlow!45, line width=.35pt]
          ({.2666667*\col},{\cy-.14}) -- ({.2666667*\col},{\cy+.14});
    }
  }
  \resizebox{\textwidth}{!}{%
    \begin{tikzpicture}[
      x=1cm, y=1cm,
      arrow/.style={draw=targetInk!80, line width=.6pt,
        -{Stealth[length=1.3mm,width=.9mm]}},
      dimension/.style={draw=targetFlow!65, line width=.45pt,
        {Stealth[length=.9mm,width=.6mm]}-{Stealth[length=.9mm,width=.6mm]}},
      operation/.style={draw=targetInk!65, fill=targetInk!4,
        rounded corners=2pt, line width=.55pt,
        minimum width=1.00cm, minimum height=1.12cm,
        font=\sffamily\fontsize{10}{11}\selectfont,
        inner sep=0pt, outer sep=0pt},
      chunk/.style={draw=targetInk!55, fill=targetInk!3,
        rounded corners=1.2pt, line width=.55pt,
        font=\ttfamily\fontsize{8}{10}\selectfont,
        minimum height=.28cm, inner xsep=3pt, inner ysep=1pt,
        anchor=west, outer sep=0pt},
      math label/.style={font=\fontsize{10}{11}\selectfont,
        inner sep=0pt, text=targetFlow},
      frame/.style={draw=targetInk!45, fill=white,
        rounded corners=2pt, line width=.55pt},
      frame title/.style={font=\sffamily\bfseries\fontsize{10}{11}\selectfont,
        text=targetInk, fill=white, anchor=west,
        inner xsep=4pt, inner ysep=1pt},
      panel label/.style={font=\sffamily\bfseries\fontsize{9}{10}\selectfont,
        inner sep=0pt, text=targetInk, anchor=east}
    ]
      \path[use as bounding box] (-.20,-.15) rectangle (16.00,5.90);
      \draw[frame] (.30,4.60) rectangle (15.95,5.70);
      \node[frame title] at (.55,5.70) {Query};
      \node[text width=15.25cm, align=left, inner sep=0pt,
        font=\sffamily\fontsize{9.5}{11}\selectfont]
        at (8.125,5.13)
        {Out of 600 \mbox{employees} in a company, 30\% got
        promoted while 10\% received bonus. How many \mbox{employees} did not
        get either a promotion or a bonus?};

      \node[panel label] at (.15,3.25) {(a)};
      \foreach \shift/\index in {0/1,8/2} {
        \begin{scope}[shift={(\shift,0)}]
          \draw[frame] (.30,2.18) rectangle (7.95,4.32);
          \node[frame title] at (.55,4.32)
            {Rationale \#\index};
          \node[operation] (pool-\index) at (4.50,3.22) {Pool};
          \node[math label] at (6.40,2.48) {$\bm z^{(\index)}$};
          \begin{scope}[shift={(5.60,3.22)}]
            \targetRows{\index}
          \end{scope}
          \draw[dimension] (5.60,3.87) -- (7.20,3.87);
          \node[math label] at (6.40,4.05) {$D$};
          \draw[dimension] (7.40,2.72) -- (7.40,3.72);
          \node[math label] at (7.69,3.22) {$S$};
        \end{scope}
      }
      \foreach \index/\cy/\trace in {
        1/3.58/{<\kern0pt<30+10=40>\kern0pt>},
        2/3.22/{<\kern0pt<100-40=60>\kern0pt>},
        3/2.86/{<\kern0pt<0.60*600=360>\kern0pt>}
      } {
        \node[chunk] (chunk-one-\index) at (.55,\cy) {\trace};
        \draw[arrow] (chunk-one-\index.east) -- (4.00,\cy);
        \draw[arrow, draw=targetFlow] (5.00,\cy) -- (5.60,\cy);
      }
      \foreach \index/\cy/\trace in {
        1/3.58/{<\kern0pt<30+10=40>\kern0pt>},
        2/3.22/{<\kern0pt<600*40/100=240>\kern0pt>},
        3/2.86/{<\kern0pt<600-240=360>\kern0pt>}
      } {
        \node[chunk] (chunk-two-\index) at (8.55,\cy) {\trace};
        \draw[arrow] (chunk-two-\index.east) -- (12.00,\cy);
        \draw[arrow, draw=targetFlow] (13.00,\cy) -- (13.60,\cy);
      }
      \node[math label, text=targetInk] at (chunk-one-3.center |- 0,2.48)
        {$r^{(1)}$};
      \node[math label, text=targetInk] at (chunk-two-2.center |- 0,2.48)
        {$r^{(2)}$};

      \draw[frame] (.30,0) rectangle (15.95,1.90);
      \node[frame title] at (.55,1.90) {Mixed Rationale};
      \node[panel label] at (.15,.95) {(b)};
      \foreach \cx/\kind in {4.10/1,8.70/2,13.10/3} {
        \begin{scope}[shift={({\cx-.80},1.04)}]
          \targetRows{\kind}
        \end{scope}
      }
      \node[math label] at (4.10,.30) {$\bm z^{(1)}$};
      \node[math label] at (8.70,.30) {$\bm z^{(2)}$};
      \node[math label] at (13.10,.30) {$\bm z_1$};
      \node[font=\fontsize{11}{12}\selectfont, inner sep=0pt, anchor=east]
        at (3.05,1.04) {$\lambda\times$};
      \node[font=\fontsize{11}{12}\selectfont, inner sep=0pt, anchor=east]
        at (7.65,1.04) {$(1-\lambda)\times$};
      \node[font=\fontsize{11}{12}\selectfont, inner sep=0pt]
        at (5.70,1.04) {$+$};
      \node[font=\fontsize{11}{12}\selectfont, inner sep=0pt]
        at (10.90,1.04) {$=$};
    \end{tikzpicture}%
  }
  \endgroup
  \caption{Rationale compression and within-query mixup.
    (a) Position-weighted embedding pooling maps each text chunk to one $D$-dimensional vector ($S=3$ shown).
    (b) A shared $\lambda$ mixes corresponding vectors of the two complete
    latent sequences. Original and mixed sequences serve as targets for
    flow matching.
    Cell colors depict schematic latent values.}
  \label{fig:flow-targets}
\end{figure*}

Despite progress in continuous reasoning, textual CoT remains a strong empirical reference~\citep{chen2026reasoning,li2025implicit}.
We thus construct a continuous counterpart to Equation~\ref{eq:cot-marginal} by explicitly distilling multiple answer-correct paths from the textual baseline into a query-conditioned flow.
The flow models a distribution $p_\theta(\bm z\mid q)$ over complete continuous rationales, allowing independent rollouts to explore alternative latent reasoning contexts for the same query.
Each sampled rationale $\bm z$ is converted into a context $\bm u$ from which the frozen ALM $p_0$ decodes an answer.
The induced context distribution $p_\theta(\bm u\mid q)$ yields the answer marginal
\begin{equation}\textstyle
p_\theta(a\mid q)
=\int p_0(a\mid\bm u,q)\,p_\theta(\bm u\mid q)\,\mathrm d\bm u.
\label{eq:continuous-answer-marginal}
\end{equation}
Thus, answer generation can vary through both rationale sampling and answer decoding, rather than decoding repeatedly from one fixed context.
We train this \emph{distributional reasoner} by distilling an empirical distribution of teacher rationales, augmented with within-query mixup.
Here we define the targets and flow objective; Section~\ref{sec:answer-aware-contextualization} describes the mapping to $\bm u$ and corresponding objectives.

\paragraph{Fixed-length rationale}
Our multi-path supervision protocol independently samples textual teacher rollouts for each query $q$ and retains $M$ answer-correct traces $\{(r^{(m)},a^{(m)})\}_{m=1}^{M}$.
We split each variable-length rationale $r^{(m)}$ into $S$ chunks.
MUX~\citep{suleymanzade2026mux} uses position-weighted distributions over vocabulary tokens as local supervision targets.
Inspired by this, we use these positional weights to average each chunk's token embeddings into a $D$-dimensional vector.
Stacking the vectors yields $\bm z^{(m)}\in\mathbb R^{S\times D}$, with one row per chunk: $S$ specifies the reasoning budget, and $D$ is the embedding dimension.
Figure~\ref{fig:flow-targets}(a) shows two such rationales for the same query with $S=3$.
We obtain token embeddings by lookup in the ALM's frozen input embedding table, so all $M$ rationale targets remain fixed throughout training.
These targets define the empirical distribution of continuous rationales for each query.

\paragraph{Within-query mixup}
The finite teacher traces provide supervision only at their observed latent targets.
Within-query mixup~\citep{zhang2018mixup,verma2019manifold} adds targets between answer-correct rationales of the same query.
For two distinct rationales from the same query, indexed by $j$ and $k$, we sample $\lambda$ and construct
\begin{equation}\textstyle
\bm z_1=\lambda\bm z^{(j)}+(1-\lambda)\bm z^{(k)}.
\label{eq:mixup-target}
\end{equation}
A shared $\lambda\sim\operatorname{Beta}(2,2)$ mixes corresponding vectors across all $S$ positions, preserving the sequence structure shown in Figure~\ref{fig:flow-targets}(b).
For each query, we train on both mixed targets and the original teacher rationale targets.
Mixup therefore augments the distribution to be distilled into the flow.
Appendix~\ref{sec:theoretical-analysis} analyzes reference-target covariance contraction, retention of original supervision, and the regularization interpretation of within-query mixup.

\paragraph{Flow matching}
We distill the original and mixed targets using flow matching~\citep{lipman2023flow,liu2023flow}.
For a target $\bm z_1$, we independently draw $\bm z_0\sim\mathcal N(\bm0,I)$ and $t\sim\mathcal U(0,1)$, then form $\bm z_t=(1-t)\bm z_0+t\bm z_1$.
We train a query-conditioned velocity field $v_\theta$ with the flow matching objective:
\begin{equation}\textstyle
\lossframe{fm}{\mathcal L_{\mathrm{flow}}=
\big\|v_\theta(\bm z_t,t,q)-(\bm z_1-\bm z_0)\big\|_F^2.}
\label{eq:flow-matching-loss}
\end{equation}

\subsection{Answer-aware contextualization}
\label{sec:answer-aware-contextualization}

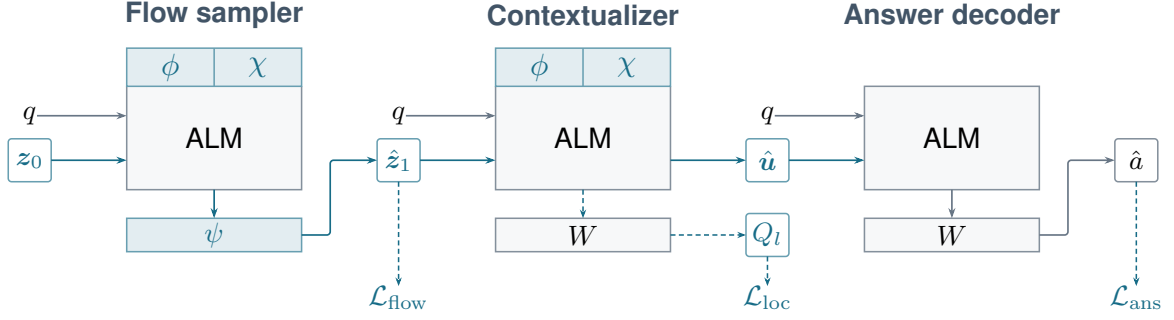
\begin{figure*}[t]
  \centering
  \begingroup
  \definecolor{architectureInk}{HTML}{425466}
  \definecolor{architectureFlow}{HTML}{176C87}
  \def\architectureModelWidth{2.20}
  \def\architectureModelHeight{1.30}
  \def\architectureCapHeight{.48}
  \def\architectureBase{.60}
  \def\architecturePortOffset{.28}
  \def\architectureSampleWidth{.54}
  \def\architectureSampleSlotWidth{1.00}
  \def\architectureGap{.72}
  \def\architectureReadoutGap{.36}
  \def\architectureReadoutHeight{.42}
  \def\architectureLossY{-.76}
  \newcommand{\architectureModel}[1]{%
    \begin{scope}[shift={(#1,\architectureBase)}]
      \fill[architectureInk!4]
        (0,0) rectangle (\architectureModelWidth,\architectureModelHeight);
      \fill[architectureFlow!12]
        (0,\architectureModelHeight) rectangle
        (\architectureModelWidth,{\architectureModelHeight+\architectureCapHeight});
      \draw[draw=architectureInk!65, line width=.55pt]
        (0,\architectureModelHeight) -- (0,0) -- (\architectureModelWidth,0)
        -- (\architectureModelWidth,\architectureModelHeight);
      \draw[draw=architectureFlow!65, line width=.55pt]
        (0,\architectureModelHeight) rectangle
        (\architectureModelWidth,{\architectureModelHeight+\architectureCapHeight});
      \draw[draw=architectureFlow!65, line width=.55pt]
        ({\architectureModelWidth/2},\architectureModelHeight)
        -- ({\architectureModelWidth/2},{\architectureModelHeight+\architectureCapHeight});
      \node[model label] at ({\architectureModelWidth/2},{\architectureModelHeight/2}) {ALM};
      \node[parameter] at ({\architectureModelWidth/4},{\architectureModelHeight+\architectureCapHeight/2}) {$\phi$};
      \node[parameter] at ({3*\architectureModelWidth/4},{\architectureModelHeight+\architectureCapHeight/2}) {$\chi$};
    \end{scope}%
  }
  \resizebox{\textwidth}{!}{%
    \begin{tikzpicture}[
      x=1cm, y=1cm,
      arrow/.style={draw=architectureInk!80, line width=.6pt,
        -{Stealth[length=1.3mm,width=.9mm]}},
      latent arrow/.style={arrow, draw=architectureFlow},
      supervision/.style={draw=architectureFlow, line width=.55pt,
        dash pattern=on 1.8pt off 1.2pt,
        -{Stealth[length=1.1mm,width=.8mm]}},
      model label/.style={font=\sffamily\fontsize{10}{11}\selectfont,
        inner sep=0pt},
      parameter/.style={font=\fontsize{11}{12}\selectfont,
        text=architectureFlow, inner sep=0pt},
      math label/.style={font=\fontsize{10}{11}\selectfont,
        inner sep=0pt},
      sample/.style={draw=architectureInk!65, fill=white,
        rounded corners=1.2pt, line width=.55pt,
        minimum height=.54cm, minimum width=\architectureSampleWidth cm,
        font=\fontsize{9.5}{11}\selectfont,
        inner xsep=0pt, inner ysep=.5pt, outer sep=0pt},
      latent sample/.style={sample, draw=architectureFlow!65,
        text=architectureFlow},
      heading/.style={font=\sffamily\bfseries\fontsize{10}{11}\selectfont,
        text=architectureInk, inner sep=0pt},
      loss/.style={font=\fontsize{10}{11}\selectfont,
        text=architectureFlow, inner sep=0pt, outer sep=0pt},
      frozen head/.style={draw=architectureInk!65, fill=architectureInk!4,
        line width=.55pt,
        minimum width=\architectureModelWidth cm,
        minimum height=\architectureReadoutHeight cm,
        font=\fontsize{10}{11}\selectfont,
        inner sep=0pt, outer sep=0pt}
    ]
      \def\noiseX{.65}
      \pgfmathsetmacro{\flowLeft}{\noiseX+\architectureSampleSlotWidth/2+\architectureGap}
      \pgfmathsetmacro{\flowRight}{\flowLeft+\architectureModelWidth}
      \pgfmathsetmacro{\rationaleX}{\flowRight+\architectureGap+\architectureSampleSlotWidth/2}
      \pgfmathsetmacro{\contextLeft}{\rationaleX+\architectureSampleSlotWidth/2+\architectureGap}
      \pgfmathsetmacro{\contextRight}{\contextLeft+\architectureModelWidth}
      \pgfmathsetmacro{\contextX}{\contextRight+\architectureGap+\architectureSampleSlotWidth/2}
      \pgfmathsetmacro{\answerLeft}{\contextX+\architectureSampleSlotWidth/2+\architectureGap}
      \pgfmathsetmacro{\answerRight}{\answerLeft+\architectureModelWidth}
      \pgfmathsetmacro{\answerX}{\answerRight+\architectureGap+\architectureSampleSlotWidth/2}
      \pgfmathsetmacro{\flowX}{(\flowLeft+\flowRight)/2}
      \pgfmathsetmacro{\contextModelX}{(\contextLeft+\contextRight)/2}
      \pgfmathsetmacro{\answerModelX}{(\answerLeft+\answerRight)/2}
      \pgfmathsetmacro{\modelCenterY}{\architectureBase+\architectureModelHeight/2}
      \pgfmathsetmacro{\architectureRow}{\modelCenterY-\architecturePortOffset}
      \pgfmathsetmacro{\queryY}{\modelCenterY+\architecturePortOffset}
      \pgfmathsetmacro{\readoutY}{\architectureBase-\architectureReadoutGap-\architectureReadoutHeight/2}
      \pgfmathsetmacro{\modelTop}{\architectureBase+\architectureModelHeight}
      \pgfmathsetmacro{\headingY}{\modelTop+\architectureCapHeight+.42}
      \path[use as bounding box] (0,{\architectureLossY-.24})
        rectangle ({\answerX+\architectureSampleSlotWidth/2+.15},{\headingY+.20});

      \node[heading] at (\flowX,\headingY) {Flow sampler};
      \node[heading] at (\contextModelX,\headingY) {Contextualizer};
      \node[heading] at (\answerModelX,\headingY) {Answer decoder};
      \architectureModel{\flowLeft}
      \architectureModel{\contextLeft}
      \fill[architectureInk!4]
        (\answerLeft,\architectureBase) rectangle
        (\answerRight,\modelTop);
      \draw[draw=architectureInk!65, line width=.55pt]
        (\answerLeft,\architectureBase) rectangle
        (\answerRight,\modelTop);
      \node[model label] at (\answerModelX,\modelCenterY) {ALM};
      \node[frozen head, draw=architectureFlow!65,
        fill=architectureFlow!12, text=architectureFlow]
        (velocityHead) at (\flowX,\readoutY) {$\psi$};
      \draw[latent arrow] (\flowX,\architectureBase) -- (velocityHead.north);
      \node[frozen head] (answerHead) at (\answerModelX,\readoutY) {$W$};
      \draw[arrow] (\answerModelX,\architectureBase) -- (answerHead.north);

      \foreach \index/\sx/\tx in {
        1/\noiseX/\flowLeft,
        2/\rationaleX/\contextLeft,
        3/\contextX/\answerLeft
      } {
        \node[math label] (query-\index) at (\sx,\queryY) {$q$};
        \draw[arrow] (query-\index.east) -- (\tx,\queryY);
      }

      \node[latent sample] (noise) at (\noiseX,\architectureRow) {$\bm z_0$};
      \node[latent sample] (rationale) at (\rationaleX,\architectureRow)
        {$\hat{\bm z}_1$};
      \node[latent sample] (context)
        at (\contextX,\architectureRow) {$\hat{\bm u}$};
      \node[sample] (answer)
        at (\answerX,\architectureRow) {$\hat a$};
      \draw[latent arrow] (noise.east) -- (\flowLeft,\architectureRow);
      \draw[latent arrow, rounded corners=1.2pt]
        (velocityHead.east) -- ({\flowRight+\architectureGap/2},\readoutY)
        -- ({\flowRight+\architectureGap/2},\architectureRow)
        -- (rationale.west);
      \draw[latent arrow] (rationale.east) -- (\contextLeft,\architectureRow);
      \draw[latent arrow] (\contextRight,\architectureRow) -- (context.west);
      \draw[latent arrow] (context.east) -- (\answerLeft,\architectureRow);
      \draw[arrow, rounded corners=1.2pt]
        (answerHead.east) -- ({\answerRight+\architectureGap/2},\readoutY)
        -- ({\answerRight+\architectureGap/2},\architectureRow)
        -- (answer.west);

      \node[loss] (flowLoss) at (\rationaleX,\architectureLossY) {$\mathcal L_{\mathrm{flow}}$};
      \draw[supervision] (rationale.south) -- (flowLoss.north);

      \node[frozen head] (localHead) at (\contextModelX,\readoutY) {$W$};
      \node[latent sample] (localPrediction)
        at (\contextX,\readoutY) {$Q_l$};
      \node[loss] (localLoss) at (\contextX,\architectureLossY)
        {$\mathcal L_{\mathrm{loc}}$};
      \draw[supervision]
        (\contextModelX,\architectureBase) -- (localHead.north);
      \draw[supervision] (localHead.east) -- (localPrediction.west);
      \draw[supervision] (localPrediction.south) -- (localLoss.north);

      \node[loss] (answerLoss) at (\answerX,\architectureLossY)
        {$\mathcal L_{\mathrm{ans}}$};
      \draw[supervision] (answer.south) -- (answerLoss.north);

    \end{tikzpicture}%
  }
  \endgroup
  \caption{Rationale sampling, contextualization, and answer decoding.
    The sampler produces $\hat{\bm z}_1$ with an ODE~solver;
    the contextualizer forms KV context $\hat{\bm u}$ at $t=1$.
    Both share $\phi$ and $\chi$, while the answer decoder generates $\hat a$
    with the reasoning adapter disabled.
    ALM weights and vocabulary projections $W$ are frozen; teal parameter
    blocks are trainable.
    Dashed arrows indicate training supervision for rationale generation,
    local content, and answer prediction.}
  \label{fig:flow-architecture}
\end{figure*}

Section~\ref{sec:rationale-distribution-distillation} defines the distribution of rationale outcomes to be distilled from the textual CoT reference.
To inherit its rationale--answer sampling structure, we must also learn how sampled continuous rationales provide contexts for answer generation.
We therefore complement flow matching with distillation of the teacher's correct answers and local reasoning content.
Figure~\ref{fig:flow-architecture} summarizes the three stages and the supervision used to train the rationale sampler and contextualizer.

\paragraph{Rationale sampling}
The velocity field $v_\theta$ uses the frozen ALM with a reasoning LoRA adapter $\phi$, a learned time embedding $\chi$, and a velocity head $\psi$, with $\theta=(\phi,\chi,\psi)$.
An ODE solver integrates this field from Gaussian noise $\bm z_0$ at $t=0$ to a complete rationale at $t=1$.
The numerical sampler $F_\theta$ maps the noise to its generated rationale:
\begin{equation}\textstyle
\hat{\bm z}_1=F_\theta(\bm z_0,q).
\label{eq:ode-sampler}
\end{equation}
Each solver evaluation predicts velocities at all $S$ positions jointly, with causal attention to the query and preceding latent positions.
Independent noise draws thus sample complete continuous reasoning paths without token-by-token rationale generation.

\paragraph{Contextualization}
A \emph{contextualizer} converts the sampled rationale into an answer context by passing $q$ and $\hat{\bm z}_1$ through the adapted ALM at $t=1$.
It shares $\phi$ and $\chi$ with the sampler but bypasses the velocity head $\psi$.
The resulting key--value (KV) caches define $\hat{\bm u}=G_{\phi,\chi}(\hat{\bm z}_1,q)$.
Applying this map to the sampled rationale distribution induces the context distribution $p_\theta(\bm u\mid q)$ in Equation~\ref{eq:continuous-answer-marginal}.

\paragraph{Answer decoding}
The frozen ALM then samples $\hat a\sim p_0(\cdot\mid\hat{\bm u},q)$ autoregressively with the reasoning adapter disabled.
It serves as both the textual teacher and the answer decoder: the teacher conditions on a textual rationale $r$, whereas the decoder conditions on the generated context $\hat{\bm u}$.
To transfer the teacher's answer predictions across these contexts, we teacher force each retained answer $a$ and minimize its sequence negative log-likelihood:
\begin{equation}\textstyle
\lossframe{ans}{\mathcal L_{\mathrm{ans}}=-\log p_0(a\mid\hat{\bm u},q).}
\label{eq:answer-loss}
\end{equation}
For a point-mass target $\delta_a$, this loss equals forward KL to the decoder distribution.
Averaging over teacher answers gives a sample-based forward-KL distillation objective, up to the fixed teacher entropy.
Because we retain only answer-correct traces, this transfer concerns the filtered empirical teacher distribution.
Gradients through the frozen decoder train the sampler and contextualizer to support the teacher's correct answers.

\paragraph{Local content distillation}
Following MUX~\citep{suleymanzade2026mux}, we supervise the contextualizer's final hidden states with local token distributions from the aligned teacher rationale chunks:
\begin{equation}\textstyle
\lossframe{loc}{\mathcal L_{\mathrm{loc}}=\sum_{l=1}^{S}\operatorname{KL}(P_l\|Q_l).}
\label{eq:local-loss}
\end{equation}
Here $P_l$ is the token distribution of chunk $l$ formed with the same positional weights as the embedding pooling, and $Q_l$ is predicted from the corresponding hidden state using the answer decoder's frozen vocabulary projection $W$ and a tempered softmax.

\paragraph{Joint objective}
For an original target, its source trace supplies the answer and local token distributions.
For a mixed target, we evaluate both source traces' answer and local losses on the same generated context and hidden states, then weight them by $\lambda$ and $1-\lambda$ as in Equation~\ref{eq:mixup-target}.
The joint objective is
\begin{equation}\textstyle
\lossframe{joint}{\mathcal L(\theta)=\mathcal L_{\mathrm{ans}}
+\lambda_{\mathrm{flow}}\mathcal L_{\mathrm{flow}}
+\lambda_{\mathrm{loc}}\mathcal L_{\mathrm{loc}}.}
\label{eq:joint-objective}
\end{equation}
Training averages over original and mixed targets, source noise, and the time samples used for flow matching.
Both auxiliary losses backpropagate through the contextualizer and ODE solver, training the context mapping and rationale sampler jointly.
Only $\theta=(\phi,\chi,\psi)$ is updated in \texttt{RDD}; pretrained ALM weights and the vocabulary projection $W$ remain fixed.
See Appendix~\ref{sec:implementation-details} for more details.

\paragraph{Inference}
At inference, each rollout starts from an independent Gaussian noise draw and follows $\bm z_0\to\hat{\bm z}_1\to\hat{\bm u}\to\hat a$, generating a fresh rationale context before decoding an answer (cf. Figure~\ref{fig:flow-architecture}).

\section{Experiments}
\label{sec:experiment}

\newlength{\hcellnumwidth}
\newcommand{\hcell}[2]{%
  \texttt{%
    \settowidth{\hcellnumwidth}{100.0}%
    \makebox[\hcellnumwidth][r]{#1}\,%
    {\footnotesize\textcolor{gray}{/\,%
      \settowidth{\hcellnumwidth}{00.0}%
      \makebox[\hcellnumwidth][r]{#2}}}%
  }%
}

\begin{table*}[t]
\centering
\setlength{\aboverulesep}{0pt}
\setlength{\belowrulesep}{0pt}
\setlength{\tabcolsep}{0.5em}
\renewcommand{\arraystretch}{1.1}
\resizebox{\textwidth}{!}{\begin{tabular}{lc*{3}{>{\centering\arraybackslash}p{2.05cm}}*{3}{>{\centering\arraybackslash}p{2.05cm}}}
  \toprule
  & & \multicolumn{3}{c}{Arithmetic benchmarks} & \multicolumn{3}{c}{Advanced benchmarks} \\
  \cmidrule(lr){3-5}\cmidrule(lr){6-8}
  Method & \# Token & \texttt{GSM8K} & \texttt{SVAMP} & \texttt{GSMHard} & \texttt{TAL} & \texttt{MATH500} & \texttt{Gaokao23} \\
  \midrule
  \texttt{CoT} & 27.4$^{\dagger}$
    & \hcell{89.5}{47.5} & \hcell{76.9}{50.8} & \hcell{23.2}{10.7}
    & \hcell{46.0}{8.1} & \hcell{43.5}{2.8} & \hcell{30.6}{1.8} \\
  \midrule
  \texttt{CoT$_{\texttt{ST}}$} & 27.2$^{\dagger}$
    & \hcell{48.5}{\UL{48.4}} & \hcell{53.3}{\BL{52.4}} & \hcell{11.4}{\BL{11.4}}
    & \hcell{10.4}{\BL{9.9}} & \hcell{4.2}{\BL{3.0}} & \hcell{3.3}{\BL{2.2}} \\
  \texttt{Coconut}$^{\ddagger}$ & 6
    & \hcell{59.7}{32.5} & \hcell{54.3}{35.2} & \hcell{14.2}{7.5}
    & \hcell{20.4}{5.1} & \hcell{13.2}{0.8} & \hcell{6.5}{0.2} \\
  \texttt{CODI} & 6
    & \hcell{76.3}{37.9} & \hcell{61.9}{40.3} & \hcell{18.6}{8.9}
    & \hcell{33.5}{6.9} & \hcell{30.7}{\UL{2.5}} & \hcell{21.1}{\UL{2.1}} \\
  \texttt{PCCoT} & 6
    & \hcell{78.2}{42.8} & \hcell{\UL{64.0}}{\UL{46.3}} & \hcell{\UL{18.8}}{9.9}
    & \hcell{\UL{38.9}}{6.6} & \hcell{35.6}{1.7} & \hcell{17.5}{0.9} \\
  \texttt{MUX} & 6
    & \hcell{\UL{78.3}}{43.4} & \hcell{62.7}{43.2} & \hcell{\UL{18.8}}{10.3}
    & \hcell{37.1}{7.0} & \hcell{\UL{38.0}}{2.2} & \hcell{\UL{28.3}}{1.2} \\
  \texttt{LOTUS} & 78
    & \hcell{75.2}{\BL{49.6}} & \hcell{62.0}{44.9} & \hcell{18.2}{\UL{11.1}}
    & \hcell{26.6}{\UL{8.4}} & \hcell{20.3}{1.6} & \hcell{14.1}{0.4} \\
  \rowcolor{blue!4}\textbf{\texttt{RDD} (ours)} & 6
    & \hcell{\BL{78.6}}{40.5} & \hcell{\BL{66.0}}{41.2} & \hcell{\BL{19.4}}{9.2}
    & \hcell{\BL{44.4}}{5.6} & \hcell{\BL{43.4}}{1.5} & \hcell{\BL{31.5}}{1.1} \\
  \bottomrule
\end{tabular}}
\caption{\textbf{Main results for GPT-2-0.1B on GSM8K-Aug.} \# Token denotes rationale length. Each result cell reports \texttt{pass@512\,{\footnotesize\textcolor{gray}{/\,pass@1}}}. Excluding \texttt{CoT}, bold-underlined and underlined values respectively denote the best and second-best results for each metric; ties share the same marking. $^{\dagger}$Averaged across all benchmarks. $^{\ddagger}$Adopted public checkpoint~\citep[\href{https://huggingface.co/connordilgren/gpt2-gsm8k-coconut}{\nolinkurl{connordilgren/gpt2-gsm8k-coconut}}; MIT license;][]{dilgren2026latent}.}
\label{tab:gpt}
\end{table*}

\begin{table*}[t]
\centering
\setlength{\aboverulesep}{0pt}
\setlength{\belowrulesep}{0pt}
\setlength{\tabcolsep}{0.5em}
\renewcommand{\arraystretch}{1.1}
\resizebox{\textwidth}{!}{\begin{tabular}{lc*{3}{>{\centering\arraybackslash}p{2.05cm}}*{3}{>{\centering\arraybackslash}p{2.05cm}}}
  \toprule
  & & \multicolumn{3}{c}{Arithmetic benchmarks} & \multicolumn{3}{c}{Advanced benchmarks} \\
  \cmidrule(lr){3-5}\cmidrule(lr){6-8}
  Method & \# Token & \texttt{GSM8K} & \texttt{SVAMP} & \texttt{GSMHard} & \texttt{TAL} & \texttt{MATH500} & \texttt{Gaokao23} \\
  \midrule
  \texttt{CoT} & 27.4$^{\dagger}$
    & \hcell{89.5}{47.5} & \hcell{76.9}{50.8} & \hcell{23.2}{10.7}
    & \hcell{46.0}{8.1} & \hcell{43.5}{2.8} & \hcell{30.6}{1.8} \\
  \midrule
  \texttt{CoT$_{\texttt{SST}}$} & 27.6$^{\dagger}$
    & \hcell{\BL{81.6}}{\BL{48.4}} & \hcell{\BL{68.3}}{\BL{51.6}} & \hcell{\BL{21.2}}{\BL{11.1}}
    & \hcell{\UL{37.1}}{\BL{8.7}} & \hcell{\UL{34.6}}{\BL{2.9}} & \hcell{\UL{27.2}}{\UL{1.9}} \\
  \texttt{CODI$_{\texttt{MCD}}$} & 6
    & \hcell{65.0}{37.7} & \hcell{56.3}{40.0} & \hcell{15.6}{8.8}
    & \hcell{27.9}{\UL{7.4}} & \hcell{20.0}{\UL{2.8}} & \hcell{16.3}{\BL{2.3}} \\
  \rowcolor{blue!4}\textbf{\texttt{RDD} (ours)} & 6
    & \hcell{\UL{78.6}}{\UL{40.5}} & \hcell{\UL{66.0}}{\UL{41.2}} & \hcell{\UL{19.4}}{\UL{9.2}}
    & \hcell{\BL{44.4}}{5.6} & \hcell{\BL{43.4}}{1.5} & \hcell{\BL{31.5}}{1.1} \\
  \bottomrule
\end{tabular}}
\caption{\textbf{Distributional rationale results for GPT-2-0.1B on GSM8K-Aug.} \# Token denotes rationale length. Each result cell reports \texttt{pass@512\,{\footnotesize\textcolor{gray}{/\,pass@1}}}. $^{\dagger}$Averaged across all benchmarks.}
\label{tab:gpt-dist}
\end{table*}

\paragraph{Models and datasets}
We use \texttt{GPT-2-0.1B}~\citep{radford2019language} and \texttt{SLM-2-1.7B}~\citep{benallal2025smollm2} as model backbones.
The training datasets are \texttt{GSM8K-Aug} and \texttt{GSM8K-Aug-NL}~\citep{deng2023implicit}, respectively.
We evaluate on six benchmarks spanning arithmetic (\texttt{GSM8K}, \texttt{SVAMP}, and \texttt{GSMHard}) and advanced mathematics (\texttt{TAL}, \texttt{MATH500}, and \texttt{Gaokao23}).
Appendix~\ref{sec:models-and-datasets} details data preparation, source references, public Hugging Face Hub releases, and reported licenses.

\paragraph{Metrics and baselines}
We report \texttt{pass@}$k$, which measures answer coverage over $k\in\{1,\dots,512\}$ independent rollouts.
While point-valued reasoning methods sample answers $a \sim p(\cdot \mid q, \bm{u})$ for a fixed context $\bm{u}$, distributional reasoning approaches sample contexts $\bm{u} \sim p(\cdot \mid q)$ as well.
Alongside \texttt{CoT}~\citep{wei2022chain} as a gold-standard reference, we compare against the following baselines: Soft Thinking~\citep[\texttt{CoT$_{\texttt{ST}}$};][]{zhang2025soft}, \texttt{Coconut}~\citep{hao2025training}, \texttt{CODI}~\citep{shen2025codi}, \texttt{PCCoT}~\citep{wu2025parallel}, \texttt{MUX}~\citep{suleymanzade2026mux}, and \texttt{LOTUS}~\citep{fan2026bridging}.
We also evaluate distributional variants: Stochastic Soft Thinking~\citep[\texttt{CoT$_{\texttt{SST}}$};][]{wu2026llms}, which uses Gumbel-Softmax sampling, and \texttt{CODI$_{\texttt{MCD}}$}~\citep{you2026parallel}, which uses Monte Carlo dropout.

\subsection{Main results}
\label{sec:main-results}

Table~\ref{tab:gpt} compares \texttt{RDD} with point-valued reasoning baselines using \texttt{GPT-2-0.1B} trained on \texttt{GSM8K-Aug}.
With a rationale budget of $S=6$, \texttt{RDD} exceeds every evaluated point-valued baseline in \texttt{pass@512} on all six benchmarks, including \texttt{Coconut}, \texttt{CODI}, \texttt{PCCoT}, and \texttt{MUX} at the same rationale length.
Notably, the improvements are more pronounced on advanced mathematics, where \texttt{RDD} approaches the textual \texttt{CoT} reference on \texttt{TAL} and \texttt{MATH500} and surpasses it on \texttt{Gaokao23}.
These results suggest that distilling textual CoT rationales into a distribution of latent rationale contexts, rather than a single fixed context, can improve answer coverage.

Table~\ref{tab:gpt-dist} further compares distributional rationale methods.
\texttt{CoT$_{\texttt{SST}}$} improves upon \texttt{CoT$_{\texttt{ST}}$} on every benchmark and leads the distributional methods on arithmetic, although it uses substantially more rationale positions on average than others.
Meanwhile, \texttt{RDD} achieves the highest coverage among these methods on all three advanced mathematics benchmarks.
It also outperforms \texttt{CODI$_{\texttt{MCD}}$} on all six benchmarks at the same budget of $S=6$.
Monte Carlo dropout reduces coverage relative to \texttt{CODI} throughout, indicating that adding inference-time stochasticity alone is insufficient to improve answer coverage in this setting.
Moreover, batched parallel inference can be non-trivial when each rollout uses independently masked model weights with Monte Carlo dropout.
In contrast, \texttt{RDD} can batch independent noise samples through a shared sampler, enabling straightforward batch-parallel inference.

\subsection{Additional results}
\label{sec:additional-results}

\begin{table}[t]
\centering
\setlength{\aboverulesep}{0pt}
\setlength{\belowrulesep}{0pt}
\setlength{\tabcolsep}{0.5em}
\renewcommand{\arraystretch}{1.1}
\settowidth{\analysislabelwidth}{\textbf{\texttt{RDD} (ours)}\hspace{1em}MPS}
\settowidth{\mpscolumnwidth}{MPS}
\resizebox{\linewidth}{!}{\begin{tabular}{p{\dimexpr\analysislabelwidth-\mpscolumnwidth-2\tabcolsep\relax}>{\centering\arraybackslash}p{\mpscolumnwidth}*{2}{>{\centering\arraybackslash}p{2.5cm}}}
  \toprule
  Method & MPS & Arithmetic & Advanced \\
  \midrule
  \texttt{CoT}
    & & \hcell{63.2}{36.3} & \hcell{40.0}{4.2} \\
  \midrule
  \texttt{CODI}
    & & \hcell{\UL{52.3}}{\UL{29.0}} & \hcell{28.4}{\BL{3.8}} \\
  \texttt{CODI}
    & $\checkmark$ & \hcell{52.0}{27.4} & \hcell{\UL{37.0}}{\UL{3.5}} \\
  \rowcolor{blue!4}\textbf{\texttt{RDD} (ours)}
    & $\checkmark$ & \hcell{\BL{54.7}}{\BL{30.3}} & \hcell{\BL{39.8}}{2.7} \\
  \bottomrule
\end{tabular}}
\caption{\textbf{Multi-path supervision results for GPT-2-0.1B on GSM8K-Aug.} MPS denotes our multi-path supervision protocol. Each cell reports \texttt{pass@512\,{\footnotesize\textcolor{gray}{/\,pass@1}}}, averaged over the three benchmarks in each group.}
\label{tab:mps}
\end{table}

\begin{table*}[t]
\centering
\setlength{\aboverulesep}{0pt}
\setlength{\belowrulesep}{0pt}
\setlength{\tabcolsep}{0.5em}
\renewcommand{\arraystretch}{1.1}
\resizebox{\textwidth}{!}{\begin{tabular}{lc*{3}{>{\centering\arraybackslash}p{2.05cm}}*{3}{>{\centering\arraybackslash}p{2.05cm}}}
  \toprule
  & & \multicolumn{3}{c}{Arithmetic benchmarks} & \multicolumn{3}{c}{Advanced benchmarks} \\
  \cmidrule(lr){3-5}\cmidrule(lr){6-8}
  Method & \# Token & \texttt{GSM8K} & \texttt{SVAMP} & \texttt{GSMHard} & \texttt{TAL} & \texttt{MATH500} & \texttt{Gaokao23} \\
  \midrule
  \texttt{CoT} & 87.9$^{\dagger}$
    & \hcell{96.7}{55.3} & \hcell{94.0}{61.9} & \hcell{46.1}{16.2}
    & \hcell{77.0}{22.4} & \hcell{69.2}{7.1} & \hcell{52.2}{5.0} \\
  \midrule
  \texttt{CoT$_{\texttt{ST}}$} & 77.9$^{\dagger}$
    & \hcell{57.5}{\BL{54.6}} & \hcell{61.7}{\UL{61.4}} & \hcell{17.4}{\BL{16.2}}
    & \hcell{28.7}{\BL{22.6}} & \hcell{18.2}{\BL{7.1}} & \hcell{10.9}{4.7} \\
  \texttt{CoT$_{\texttt{SST}}$} & 78.8$^{\dagger}$
    & \hcell{\BL{94.2}}{\UL{54.2}} & \hcell{\BL{91.7}}{\BL{62.0}} & \hcell{\BL{42.8}}{\UL{16.0}}
    & \hcell{\BL{74.6}}{\UL{22.0}} & \hcell{\BL{66.4}}{\UL{6.9}} & \hcell{\BL{52.2}}{\BL{5.1}} \\
  \texttt{CODI} & 6
    & \hcell{84.8}{27.4} & \hcell{77.3}{38.9} & \hcell{22.7}{6.6}
    & \hcell{61.0}{12.0} & \hcell{52.0}{3.6} & \hcell{33.7}{3.9} \\
  \texttt{MUX} & 6
    & \hcell{\UL{87.6}}{27.6} & \hcell{76.0}{39.1} & \hcell{\UL{24.3}}{6.9}
    & \hcell{62.0}{13.1} & \hcell{55.0}{5.0} & \hcell{32.6}{\UL{4.8}} \\
  \rowcolor{blue!4}\textbf{\texttt{RDD} (ours)} & 6
    & \hcell{\UL{87.6}}{21.2} & \hcell{\UL{80.7}}{37.2} & \hcell{\UL{24.3}}{5.1}
    & \hcell{\UL{63.3}}{11.6} & \hcell{\UL{56.0}}{3.9} & \hcell{\UL{35.9}}{3.8} \\
  \bottomrule
\end{tabular}}
\caption{\textbf{Additional results for SLM-2-1.7B on GSM8K-Aug-NL.} \# Token denotes rationale length. Each result cell reports \texttt{pass@512\,{\footnotesize\textcolor{gray}{/\,pass@1}}}. Excluding \texttt{CoT}, bold-underlined and underlined values respectively denote the best and second-best results for each metric; ties share the same marking. $^{\dagger}$Averaged across all benchmarks.}
\label{tab:slm}
\end{table*}

\paragraph{Multi-path supervision}
Our multi-path supervision protocol uses multiple answer-correct textual CoTs per query as teacher targets.
We apply this protocol to \texttt{CODI} to examine whether even a point-valued reasoner can benefit from multiple teacher paths.
In Table~\ref{tab:mps}, we observe a slight decrease in \texttt{CODI}'s mean \texttt{pass@512} on arithmetic, from 52.3\% to 52.0\%, consistent with prior findings~\citep{suleymanzade2026limits}, but a substantial improvement on advanced mathematics, from 28.4\% to 37.0\%.
With the same supervision, \texttt{RDD} learns a distribution over rationale contexts and improves on \texttt{CODI} in both benchmark groups, reaching 54.7\% on arithmetic and 39.8\% on advanced mathematics.
These findings support the value of our multi-path supervision on advanced mathematics and suggest that distributional rationale learning can make more effective use of the same teacher paths.

\begin{figure}[t]
  \centering
  \includegraphics[width=\linewidth]{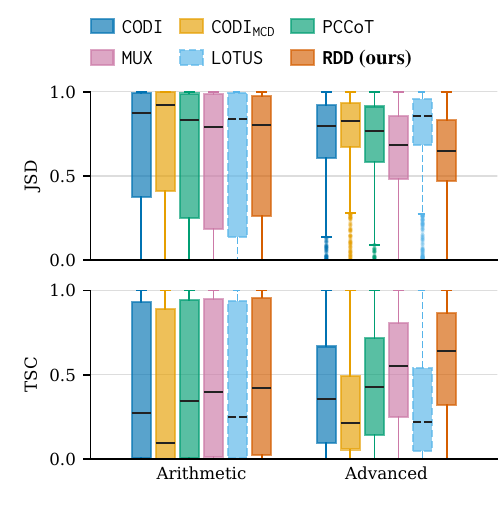}
  \caption{\textbf{Answer-distribution fidelity to textual CoT.}
  Query-level JSD ($\downarrow$) and TSC ($\uparrow$) pooled by group; see Figure~\ref{fig:fidelity-benchmarks} for per-benchmark results.}
  \label{fig:fidelity-analysis}
\end{figure}

\paragraph{Fidelity analysis}
\label{par:fidelity-analysis}
For each query, we compare empirical answer distributions estimated from sampled 512 rollouts of textual \texttt{CoT} and each latent method using Jensen--Shannon divergence (JSD) and teacher support coverage (TSC), both defined in Appendix~\ref{sec:evaluation}.
These metrics assess how closely $p_\theta(a\mid q)$ matches $p_{\mathrm{CoT}}(a\mid q)$, defined in Equations~\ref{eq:continuous-answer-marginal} and~\ref{eq:cot-marginal}, respectively.
Figure~\ref{fig:fidelity-analysis} pools question-level measurements within the arithmetic and advanced benchmark groups; Appendix~\ref{sec:supplementary-results} provides the per-benchmark breakdown.
\texttt{RDD} achieves the lowest JSD and the highest TSC on average.
It implies that \texttt{RDD} better matches the teacher's answer distribution, especially on advanced tasks, complementing its answer coverage advantage.

\paragraph{SLM-2-1.7B on GSM8K-Aug-NL}
Our comprehensive main comparison uses \texttt{GPT-2-0.1B} on \texttt{GSM8K-Aug}, a common evaluation setup for latent reasoning.
To assess whether the findings extend to a larger backbone and more elaborate natural-language rationales rather than compact symbolic derivations, we additionally evaluate \texttt{SLM-2-1.7B} trained on \texttt{GSM8K-Aug-NL} in Table~\ref{tab:slm}, where \texttt{CODI} is trained with MPS.
\texttt{RDD} continues to outperform the evaluated latent baselines at the same rationale budget of $S=6$.
However, the gap to textual \texttt{CoT} and \texttt{CoT$_{\texttt{SST}}$} is generally larger in this setting.
The longer natural-language rationales impose substantially stronger compression into six latent positions, which may contribute to this gap.
Despite this demanding compression regime, \texttt{RDD} retains its advantage in answer coverage.

\paragraph{Trajectory-level mixup compatibility}
\label{par:trajectory-level-mixup-compatibility}
We assess within-query interpolation of uncompressed rationale embeddings with \texttt{CoT}.
We test midpoint interpolation ($\lambda=1/2$) on 10,000 triples $(q,r_1,r_2)$, where both rationales are answer-correct for $q$ and have matching numbers of reasoning steps and token counts at each step.
With query embeddings fixed, \texttt{CoT} achieves 100\% accuracy using $r_1$, 99.99\% with within-query interpolation, and 44.33\% when $r_2$ is replaced by a rationale from another query.
This probe supports answer compatibility in an aligned, uncompressed setting; Appendix~\ref{sec:mixup-empirical-evidence} discusses mixup regularization.

\subsection{Ablation study}
\label{sec:ablation-study}

\begin{table}[t]
\centering
\setlength{\aboverulesep}{0pt}
\setlength{\belowrulesep}{0pt}
\setlength{\tabcolsep}{0.5em}
\renewcommand{\arraystretch}{1.1}
\settowidth{\analysislabelwidth}{\textbf{\texttt{RDD} (ours)}\hspace{1em}MPS}
\resizebox{\linewidth}{!}{\begin{tabular}{p{\analysislabelwidth}*{2}{>{\centering\arraybackslash}p{2.5cm}}}
  \toprule
  Ablation & Arithmetic & Advanced \\
  \midrule
  \rowcolor{black!4}\multicolumn{3}{c}{\textbf{\textit{Per-rollout rationale sampling}}} \\
  Fixed
    & \hcell{54.0}{30.5} & \hcell{38.5}{2.7} \\
  \rowcolor{blue!4}Sampled
    & \hcell{54.7}{30.3} & \hcell{39.8}{2.7} \\
  \midrule
  \rowcolor{black!4}\multicolumn{3}{c}{\textbf{\textit{Number of function evaluations}}} \\
  NFE=3
    & \hcell{54.6}{22.7} & \hcell{40.7}{3.0} \\
  \rowcolor{blue!4}NFE=6
    & \hcell{54.7}{30.3} & \hcell{39.8}{2.7} \\
  \midrule
  \rowcolor{black!4}\multicolumn{3}{c}{\textbf{\textit{Mixing-weight distribution}}} \\
  No-mixing
    & \hcell{52.6}{25.4} & \hcell{36.2}{3.0} \\
  \rowcolor{blue!4}$\lambda \sim \operatorname{Beta}(2,2)$
    & \hcell{54.7}{30.3} & \hcell{39.8}{2.7} \\
  $\lambda \sim \operatorname{Beta}(5,5)$
    & \hcell{53.9}{23.2} & \hcell{40.8}{2.9} \\
  $\lambda = 1/2$
    & \hcell{52.5}{32.6} & \hcell{38.6}{2.8} \\
  \midrule
  \rowcolor{black!4}\multicolumn{3}{c}{\textbf{\textit{Causal versus bidirectional attention}}} \\
  Bidirectional
    & \hcell{52.1}{24.6} & \hcell{38.5}{2.5} \\
  \rowcolor{blue!4}Causal
    & \hcell{54.7}{30.3} & \hcell{39.8}{2.7} \\
  \bottomrule
\end{tabular}}
\caption{\textbf{Ablation results for GPT-2-0.1B.} Each cell reports \texttt{pass@512\,{\footnotesize\textcolor{gray}{/\,pass@1}}}, averaged over the three benchmarks in each group.}
\label{tab:ablation}
\end{table}

\paragraph{Per-rollout rationale sampling}
\label{par:rationale-resampling}
We evaluate the contribution of sampling a fresh rationale context for each rollout using the same trained \texttt{RDD} checkpoint.
The fixed-context control retains the distributionally trained model but reuses a single rationale context sampled from its learned distribution for each query.
Both conditions use 512 answer rollouts per query and identical answer-decoding settings.
Table~\ref{tab:ablation} shows that sampling a fresh context for each rollout improves mean \texttt{pass@512} in both benchmark groups.
These results suggest that sampling multiple rationale contexts helps translate the learned distribution into broader answer coverage.

\paragraph{Number of function evaluations}
\label{par:number-of-function-evaluations}
We compare three and six Euler steps, matching the number of function evaluations (NFE) during training and inference for each model.
Table~\ref{tab:ablation} shows that NFE=3 yields comparable \texttt{pass@512} in both benchmark groups while halving flow evaluations.
At $S=6$, this also reduces sequential model forward passes relative to autoregressive baselines, which generate a full latent rationale token by token in $S$ passes.

\paragraph{Mixing-weight distribution}
\label{par:mixing-weight-distribution}
We compare no mixing ($\lambda=1$), stochastic mixing with Beta-distributed weights, and deterministic midpoint mixing ($\lambda=1/2$).
Table~\ref{tab:ablation} shows that stochastic mixing improves coverage over no mixing in both benchmark groups, especially on advanced tasks.
Midpoint mixing improves only advanced coverage.
These gains support a regularization interpretation of within-query mixup; Appendix~\ref{sec:theoretical-analysis} explains why reference-target covariance contraction alone does not determine answer coverage.

\paragraph{Causal versus bidirectional attention}
\label{par:causal-versus-bidirectional-attention}
Early work on adapting pretrained ALMs to diffusion language models advocated moving from causal to bidirectional attention~\citep{gong2025scaling}.
More recent work, however, finds that retaining causal attention can yield better reasoning performance when adapting pretrained ALMs~\citep{ma2026ar}.
Since these findings suggest no universally preferred attention pattern, we compare causal and bidirectional attention in \texttt{RDD}, which also reuses pretrained ALM weights.
Table~\ref{tab:ablation} shows that retaining causal attention yields better performance in our experiments, consistent with the latter finding.

\FloatBarrier
\section{Conclusion}
\label{sec:conclusion}

We introduced Rationale Distribution Distillation (RDD), a flow-matching approach to transferring multiple textual reasoning paths into a distribution of continuous latent rationales.
Rather than relying on a single latent trajectory to implicitly capture reasoning alternatives, RDD explicitly learns their query-conditioned distribution from answer-correct textual CoT traces, augmented through within-query mixup.
It enables exploration of multiple plausible rationales, while parallel refinement of fixed-length latent sequences replaces token-by-token rationale generation.
Our main experimental results show that RDD improves answer coverage over the evaluated point-valued latent reasoning baselines across arithmetic and advanced mathematical tasks, even nearly matching the textual CoT reference on the advanced benchmarks.
These findings highlight distributional modeling as a promising direction for transferring the benefits of multi-path textual reasoning into compact latent representations.
An intriguing avenue for future research is developing more effective rationale compression and sampling strategies to extend distributional reasoning to larger scales.

\section{Limitations}
Our experiments use backbones up to 1.7B parameters and cover both short symbolic rationales and natural-language rationales with verbal explanations, building on evaluation settings established in the latent reasoning literature~\citep{shen2025codi}.
However, these rationales remain substantially shorter and less elaborate than the extended reasoning traces produced by modern large language models~\citep{zhang2025days}.
Whether the benefits of rationale distribution distillation extend to substantially larger models and longer, more verbal reasoning traces remains an open question.

Moreover, although \texttt{RDD} improves answer coverage (\texttt{pass@$k$}), its single-rollout accuracy (\texttt{pass@1}) generally trails the strongest baselines.
Our experiments focus on supervised distillation, leaving open whether reward-based optimization can narrow this gap.
\citet{wu2026llms} identify stochastic continuous reasoning as a foundation for reinforcement learning (RL), and \citet{tang2026multiplex} demonstrate RL optimization of stochastic token-wise branch-and-merge trajectories.
These findings motivate future RL optimization of \texttt{RDD}'s sampled latent rationales to improve \texttt{pass@1} while retaining \texttt{pass@$k$}.

\section{Ethical Concerns}
Continuous latent reasoning raises concerns about transparency and oversight because its intermediate states are not directly interpretable in natural language.
Although textual reasoning traces can also be unfaithful to model behavior~\citep{chen2025reasoning}, they can at least support chain-of-thought monitoring~\citep{korbak2025chain}.
Reasoning in latent space may reduce access to these signals, motivating oversight methods for latent reasoning.

\bibliographystyle{plainnat}
\bibliography{main}

@article{guo2025deepseek-r1,
  author        = {Guo, D. and Yang, D. and Zhang, H. and Song, J. and Wang, P. and Zhu, Q. and Xu, R. and Zhang, R. and Ma, S. and Bi, X. and Zhang, X. and Yu, X. and Wu, Y. and Wu, Z. F. and Gou, Z. and Shao, Z. and Li, Z. and Gao, Z. and Liu, A. and Xue, B. and Wang, B. and Wu, B. and Feng, B. and Lu, C. and Zhao, C. and Deng, C. and Ruan, C. and Dai, D. and Chen, D. and Ji, D. and Li, E. and Lin, F. and Dai, F. and Luo, F. and Hao, G. and Chen, G. and Li, G. and Zhang, H. and Xu, H. and Ding, H. and Gao, H. and Qu, H. and Li, H. and Guo, J. and Li, J. and Chen, J. and Yuan, J. and Tu, J. and Qiu, J. and Li, J. and Cai, J. L. and Ni, J. and Liang, J. and Chen, J. and Dong, K. and Hu, K. and You, K. and Gao, K. and Guan, K. and Huang, K. and Yu, K. and Wang, L. and Zhang, L. and Zhao, L. and Wang, L. and Zhang, L. and Xu, L. and Xia, L. and Zhang, M. and Zhang, M. and Tang, M. and Zhou, M. and Li, M. and Wang, M. and Li, M. and Tian, N. and Huang, P. and Zhang, P. and Wang, Q. and Chen, Q. and Du, Q. and Ge, R. and Zhang, R. and Pan, R. and Wang, R. and Chen, R. J. and Jin, R. L. and Chen, R. and Lu, S. and Zhou, S. and Chen, S. and Ye, S. and Wang, S. and Yu, S. and Zhou, S. and Pan, S. and Li, S. S. and Zhou, S. and Wu, S. and Yun, T. and Pei, T. and Sun, T. and Wang, T. and Zeng, W. and Liu, W. and Liang, W. and Gao, W. and Yu, W. and Zhang, W. and Xiao, W. L. and An, W. and Liu, X. and Wang, X. and Chen, X. and Nie, X. and Cheng, X. and Liu, X. and Xie, X. and Liu, X. and Yang, X. and Li, X. and Su, X. and Lin, X. and Li, X. Q. and Jin, X. and Shen, X. and Chen, X. and Sun, X. and Wang, X. and Song, X. and Zhou, X. and Wang, X. and Shan, X. and Li, Y. K. and Wang, Y. Q. and Wei, Y. X. and Zhang, Y. and Xu, Y. and Li, Y. and Zhao, Y. and Sun, Y. and Wang, Y. and Yu, Y. and Zhang, Y. and Shi, Y. and Xiong, Y. and He, Y. and Piao, Y. and Wang, Y. and Tan, Y. and Ma, Y. and Liu, Y. and Guo, Y. and Ou, Y. and Wang, Y. and Gong, Y. and Zou, Y. and He, Y. and Xiong, Y. and Luo, Y. and You, Y. and Liu, Y. and Zhou, Y. and Zhu, Y. X. and Huang, Y. and Li, Y. and Zheng, Y. and Zhu, Y. and Ma, Y. and Tang, Y. and Zha, Y. and Yan, Y. and Ren, Z. Z. and Ren, Z. and Sha, Z. and Fu, Z. and Xu, Z. and Xie, Z. and Zhang, Z. and Hao, Z. and Ma, Z. and Yan, Z. and Wu, Z. and Gu, Z. and Zhu, Z. and Liu, Z. and Li, Z. and Xie, Z. and Song, Z. and Pan, Z. and Huang, Z. and Xu, Z. and Zhang, Z. and Zhang, Z.},
  title         = {{DeepSeek-R1} incentivizes reasoning in {LLMs} through reinforcement learning},
  journal       = {Nature},
  year          = {2025},
  volume        = {645},
  pages         = {633--638},
  doi           = {10.1038/s41586-025-09422-z}
}

@article{cui2022accelerating,
  author        = {Cui, Y. and Yao, W. and Li, Q. and Chan, A. B. and Xue, C. J.},
  title         = {Accelerating {Monte Carlo Bayesian} Prediction via Approximating Predictive Uncertainty Over the Simplex},
  journal       = {IEEE Transactions on Neural Networks and Learning Systems},
  year          = {2022},
  volume        = {33},
  number        = {4},
  pages         = {1492--1506},
  doi           = {10.1109/TNNLS.2020.3042525}
}

@article{carratino2022mixup,
  author        = {Carratino, L. and Ciss{\'e}, M. and Jenatton, R. and Vert, J.-P.},
  title         = {On Mixup Regularization},
  journal       = {Journal of Machine Learning Research},
  year          = {2022},
  volume        = {23},
  number        = {325},
  pages         = {1--31},
  url           = {https://www.jmlr.org/papers/v23/20-1385.html}
}

@article{lin1991divergence,
  author        = {Lin, J.},
  title         = {Divergence Measures Based on the {Shannon} Entropy},
  journal       = {IEEE Transactions on Information Theory},
  year          = {1991},
  volume        = {37},
  number        = {1},
  pages         = {145--151},
  doi           = {10.1109/18.61115}
}

@article{gibbs2002choosing,
  author        = {Gibbs, A. L. and Su, F. E.},
  title         = {On Choosing and Bounding Probability Metrics},
  journal       = {International Statistical Review},
  year          = {2002},
  volume        = {70},
  number        = {3},
  pages         = {419--435},
  doi           = {10.1111/j.1751-5823.2002.tb00178.x}
}

@inproceedings{hao2025training,
  author        = {Hao, S. and Sukhbaatar, S. and Su, D. and Li, X. and Hu, Z. and Weston, J. E. and Tian, Y.},
  title         = {Training large language models to reason in a continuous latent space},
  booktitle     = {Conference on Language Modeling (COLM)},
  year          = {2025}
}

@inproceedings{wei2022chain,
  author        = {Wei, J. and Wang, X. and Schuurmans, D. and Bosma, M. and Ichter, B. and Xia, F. and Chi, E. H. and Le, Q. V. and Zhou, D.},
  title         = {Chain-of-thought prompting elicits reasoning in large language models},
  booktitle     = {Conference on Neural Information Processing Systems (NeurIPS)},
  year          = {2022}
}

@inproceedings{chen2025do,
  author        = {Chen, X. and Xu, J. and Liang, T. and He, Z. and Pang, J. and Yu, D. and Song, L. and Liu, Q. and Zhou, M. and Zhang, Z. and Wang, R. and Tu, Z. and Mi, H. and Yu, D.},
  title         = {Do {NOT} think that much for 2+3=? on the overthinking of long reasoning models},
  booktitle     = {International Conference on Machine Learning (ICML)},
  year          = {2025}
}

@inproceedings{han2025token,
  author        = {Han, T. and Wang, Z. and Fang, C. and Zhao, S. and Ma, S. and Chen, Z.},
  title         = {Token-budget-aware {LLM} reasoning},
  booktitle     = {Findings of the Association for Computational Linguistics: ACL 2025},
  year          = {2025}
}

@inproceedings{ma2025cot,
  author        = {Ma, X. and Wan, G. and Yu, R. and Fang, G. and Wang, X.},
  title         = {{CoT-Valve}: Length-compressible chain-of-thought tuning},
  booktitle     = {Proceedings of the 63rd Annual Meeting of the Association for Computational Linguistics (Volume 1: Long Papers)},
  year          = {2025}
}

@inproceedings{munkhbat2025self,
  author        = {Munkhbat, T. and Ho, N. and Kim, S. H. and Yang, Y. and Kim, Y. and Yun, S.-Y.},
  title         = {Self-training elicits concise reasoning in large language models},
  booktitle     = {Findings of the Association for Computational Linguistics: ACL 2025},
  year          = {2025}
}

@inproceedings{kang2025c3ot,
  author        = {Kang, Y. and Sun, X. and Chen, L. and Zou, W.},
  title         = {{C3oT}: Generating shorter chain-of-thought without compromising effectiveness},
  booktitle     = {AAAI Conference on Artificial Intelligence},
  year          = {2025}
}

@inproceedings{zhu2025reasoning,
  author        = {Zhu, H. and Hao, S. and Hu, Z. and Jiao, J. and Russell, S. and Tian, Y.},
  title         = {Reasoning by superposition: A theoretical perspective on chain of continuous thought},
  booktitle     = {Conference on Neural Information Processing Systems (NeurIPS)},
  year          = {2025}
}

@inproceedings{gozeten2026continuous,
  author        = {Gozeten, H. A. and Ildiz, M. E. and Zhang, X. and Harutyunyan, H. and Rawat, A. S. and Oymak, S.},
  title         = {Continuous chain of thought enables parallel exploration and reasoning},
  booktitle     = {International Conference on Learning Representations (ICLR)},
  year          = {2026}
}

@inproceedings{zhang2025soft,
  author        = {Zhang, Z. and He, X. and Yan, W. and Shen, A. and Zhao, C. and Wang, X. E.},
  title         = {Soft thinking: Unlocking the reasoning potential of {LLMs} in continuous concept space},
  booktitle     = {Conference on Neural Information Processing Systems (NeurIPS)},
  year          = {2025}
}

@inproceedings{suleymanzade2026limits,
  author        = {Suleymanzade, A. and Bergmeister, A. and Jegelka, S.},
  title         = {Limits of Continuous Chain-of-Thought in Multi-Step and Multi-Chain Reasoning},
  booktitle     = {Logical and Symbolic Reasoning in Language Models @ AAAI 2026},
  year          = {2026}
}

@inproceedings{goyal2024think,
  author        = {Goyal, S. and Ji, Z. and Rawat, A. S. and Menon, A. K. and Kumar, S. and Nagarajan, V.},
  title         = {Think before you speak: Training language models with pause tokens},
  booktitle     = {International Conference on Learning Representations (ICLR)},
  year          = {2024}
}

@inproceedings{giannou2023looped,
  author        = {Giannou, A. and Rajput, S. and Sohn, J.-Y. and Lee, K. and Lee, J. D. and Papailiopoulos, D.},
  title         = {Looped transformers as programmable computers},
  booktitle     = {International Conference on Machine Learning (ICML)},
  year          = {2023}
}

@inproceedings{gat2024discrete,
  author        = {Gat, I. and Remez, T. and Shaul, N. and Kreuk, F. and Chen, R. T. Q. and Synnaeve, G. and Adi, Y. and Lipman, Y.},
  title         = {Discrete flow matching},
  booktitle     = {Conference on Neural Information Processing Systems (NeurIPS)},
  year          = {2024}
}

@inproceedings{shen2025codi,
  author        = {Shen, Z. and Yan, H. and Zhang, L. and Hu, Z. and Du, Y. and He, Y.},
  title         = {{CODI}: Compressing {Chain-of-Thought} into continuous space via self-distillation},
  booktitle     = {Proceedings of the 2025 Conference on Empirical Methods in Natural Language Processing},
  year          = {2025}
}

@inproceedings{wu2025parallel,
  author        = {Wu, H. and Teng, Z. and Tu, K.},
  title         = {Parallel continuous {Chain-of-Thought} with {Jacobi} iteration},
  booktitle     = {Proceedings of the 2025 Conference on Empirical Methods in Natural Language Processing},
  year          = {2025}
}

@inproceedings{wei2026sim,
  author        = {Wei, X. and Liu, X. and Zang, Y. and Dong, X. and Cao, Y. and Wang, J. and Qiu, X. and Lin, D.},
  title         = {{SIM-CoT}: Supervised implicit {Chain-of-Thought}},
  booktitle     = {International Conference on Learning Representations (ICLR)},
  year          = {2026}
}

@inproceedings{tan2025think,
  author        = {Tan, W. and Li, J. and Ju, J. and Luo, Z. and Song, R. and Luan, J.},
  title         = {Think silently, think fast: Dynamic latent compression of {LLM} reasoning chains},
  booktitle     = {Conference on Neural Information Processing Systems (NeurIPS)},
  year          = {2025}
}

@inproceedings{wu2026llms,
  author        = {Wu, J. and Lu, J. and Ren, Z. and Hu, G. and Wu, Z. and Dai, D. and Wu, H.},
  title         = {{LLMs} are single-threaded reasoners: Demystifying the working mechanism of soft thinking},
  booktitle     = {International Conference on Learning Representations (ICLR)},
  year          = {2026}
}

@inproceedings{kang2026ladir,
  author        = {Kang, H. and Zhang, Y. and Kuang, N. L. and Majamaki, N. and Jaitly, N. and Ma, Y.-A. and Qin, L.},
  title         = {{LaDiR}: Latent diffusion enhances {LLMs} for text reasoning},
  booktitle     = {International Conference on Learning Representations (ICLR)},
  year          = {2026}
}

@inproceedings{you2026parallel,
  author        = {You, R. and Li, Y. and Liu, M. and Wang, W. and Nie, L. and Li, W.},
  title         = {Parallel test-time scaling for latent reasoning models},
  booktitle     = {Proceedings of the 64th Annual Meeting of the Association for Computational Linguistics (Volume 1: Long Papers)},
  year          = {2026}
}

@inproceedings{zhang2018mixup,
  author        = {Zhang, H. and Cisse, M. and Dauphin, Y. N. and Lopez-Paz, D.},
  title         = {mixup: Beyond empirical risk minimization},
  booktitle     = {International Conference on Learning Representations (ICLR)},
  year          = {2018}
}

@inproceedings{lipman2023flow,
  author        = {Lipman, Y. and Chen, R. T. Q. and Ben-Hamu, H. and Nickel, M. and Le, M.},
  title         = {Flow matching for generative modeling},
  booktitle     = {International Conference on Learning Representations (ICLR)},
  year          = {2023}
}

@inproceedings{liu2023flow,
  author        = {Liu, X. and Gong, C. and Liu, Q.},
  title         = {Flow Straight and Fast: Learning to Generate and Transfer Data with Rectified Flow},
  booktitle     = {International Conference on Learning Representations (ICLR)},
  year          = {2023}
}

@inproceedings{ling2017program,
  author        = {Ling, W. and Yogatama, D. and Dyer, C. and Blunsom, P.},
  title         = {Program Induction by Rationale Generation: Learning to Solve and Explain Algebraic Word Problems},
  booktitle     = {Proceedings of the 55th Annual Meeting of the Association for Computational Linguistics (Volume 1: Long Papers)},
  year          = {2017}
}

@inproceedings{kojima2022large,
  author        = {Kojima, T. and Gu, S. and Reid, M. and Matsuo, Y. and Iwasawa, Y.},
  title         = {Large Language Models are Zero-Shot Reasoners},
  booktitle     = {Conference on Neural Information Processing Systems (NeurIPS)},
  year          = {2022}
}

@inproceedings{patel2021nlp,
  author        = {Patel, A. and Bhattamishra, S. and Goyal, N.},
  title         = {Are {NLP} Models really able to Solve Simple Math Word Problems?},
  booktitle     = {Proceedings of the 2021 Conference of the North American Chapter of the Association for Computational Linguistics: Human Language Technologies},
  year          = {2021}
}

@inproceedings{hendrycks2021measuring,
  author        = {Hendrycks, D. and Burns, C. and Kadavath, S. and Arora, A. and Basart, S. and Tang, E. and Song, D. and Steinhardt, J.},
  title         = {Measuring Mathematical Problem Solving With the {MATH} Dataset},
  booktitle     = {NeurIPS Datasets and Benchmarks},
  year          = {2021}
}

@inproceedings{malinin2020ensemble,
  author        = {Malinin, A. and Mlodozeniec, B. and Gales, M.},
  title         = {Ensemble Distribution Distillation},
  booktitle     = {International Conference on Learning Representations (ICLR)},
  year          = {2020}
}

@inproceedings{park2025ensemble,
  author        = {Park, J. and Nam, G. and Kim, H. and Yoon, J. and Lee, J.},
  title         = {Ensemble Distribution Distillation via Flow Matching},
  booktitle     = {International Conference on Machine Learning (ICML)},
  year          = {2025}
}

@inproceedings{li2023symbolic,
  author        = {Li, L. H. and Hessel, J. and Yu, Y. and Ren, X. and Chang, K.-W. and Choi, Y.},
  title         = {Symbolic {Chain-of-Thought} Distillation: Small Models Can Also ``Think'' Step-by-Step},
  booktitle     = {Proceedings of the 61st Annual Meeting of the Association for Computational Linguistics (Volume 1: Long Papers)},
  year          = {2023}
}

@inproceedings{chen2023mcc,
  author        = {Chen, H. and Wu, S. and Quan, X. and Wang, R. and Yan, M. and Zhang, J.},
  title         = {{MCC-KD}: {Multi-CoT} Consistent Knowledge Distillation},
  booktitle     = {Findings of the Association for Computational Linguistics: EMNLP 2023},
  year          = {2023}
}

@inproceedings{ho2023large,
  author        = {Ho, N. and Schmid, L. and Yun, S.-Y.},
  title         = {Large Language Models Are Reasoning Teachers},
  booktitle     = {Proceedings of the 61st Annual Meeting of the Association for Computational Linguistics (Volume 1: Long Papers)},
  year          = {2023}
}

@inproceedings{hu2024flow,
  author        = {Hu, V. T. and Wu, D. and Asano, Y. M. and Mettes, P. and Fernando, B. and Ommer, B. and Snoek, C. G. M.},
  title         = {Flow Matching for Conditional Text Generation in a Few Sampling Steps},
  booktitle     = {Proceedings of the 18th Conference of the European Chapter of the Association for Computational Linguistics (Volume 2: Short Papers)},
  year          = {2024}
}

@inproceedings{xu2025softcot,
  author        = {Xu, Y. and Guo, X. and Zeng, Z. and Miao, C.},
  title         = {{SoftCoT}: Soft {Chain-of-Thought} for Efficient Reasoning with {LLMs}},
  booktitle     = {Proceedings of the 63rd Annual Meeting of the Association for Computational Linguistics (Volume 1: Long Papers)},
  year          = {2025}
}

@inproceedings{liu2025deliberation,
  author        = {Liu, L. and Pfeiffer, J. and Wu, J. and Xie, J. and Szlam, A.},
  title         = {Deliberation in Latent Space via Differentiable Cache Augmentation},
  booktitle     = {International Conference on Machine Learning (ICML)},
  year          = {2025}
}

@inproceedings{verma2019manifold,
  author        = {Verma, V. and Lamb, A. and Beckham, C. and Najafi, A. and Mitliagkas, I. and Lopez-Paz, D. and Bengio, Y.},
  title         = {Manifold Mixup: Better Representations by Interpolating Hidden States},
  booktitle     = {International Conference on Machine Learning (ICML)},
  year          = {2019}
}

@inproceedings{lee2025generalized,
  author        = {Lee, C. and Im, J. and Kim, J. H. T.},
  title         = {A Generalized Theory of Mixup for Structure-Preserving Synthetic Data},
  booktitle     = {International Conference on Artificial Intelligence and Statistics (AISTATS)},
  year          = {2025}
}

@inproceedings{dilgren2026latent,
  author        = {Dilgren, C. and Wiegreffe, S.},
  title         = {Are Latent Reasoning Models Easily Interpretable?},
  booktitle     = {Conference on Language Modeling (COLM)},
  year          = {2026}
}

@inproceedings{gong2025scaling,
  author        = {Gong, S. and Agarwal, S. and Zhang, Y. and Ye, J. and Zheng, L. and Li, M. and An, C. and Zhao, P. and Bi, W. and Han, J. and Peng, H. and Kong, L.},
  title         = {Scaling Diffusion Language Models via Adaptation from Autoregressive Models},
  booktitle     = {International Conference on Learning Representations (ICLR)},
  year          = {2025}
}

@inproceedings{ma2026ar,
  author        = {Ma, X. and Xiao, T. and Li, Z. and Zhang, L.},
  title         = {From {AR} to Diffusion: Efficiently Adapting Large Language Models with Strictly Causal and Elastic Horizons},
  booktitle     = {Proceedings of the 64th Annual Meeting of the Association for Computational Linguistics (Volume 1: Long Papers)},
  year          = {2026}
}

@inproceedings{hu2022lora,
  author        = {Hu, E. J. and Shen, Y. and Wallis, P. and Allen-Zhu, Z. and Li, Y. and Wang, S. and Wang, L. and Chen, W.},
  title         = {{LoRA}: Low-rank adaptation of large language models},
  booktitle     = {International Conference on Learning Representations (ICLR)},
  year          = {2022}
}

@inproceedings{hayou2024loraplus,
  author        = {Hayou, S. and Ghosh, N. and Yu, B.},
  title         = {{LoRA}+: Efficient low rank adaptation of large models},
  booktitle     = {International Conference on Machine Learning (ICML)},
  year          = {2024}
}

@inproceedings{loshchilov2019decoupled,
  author        = {Loshchilov, I. and Hutter, F.},
  title         = {Decoupled Weight Decay Regularization},
  booktitle     = {International Conference on Learning Representations (ICLR)},
  year          = {2019}
}

@inproceedings{hagele2024scaling,
  author        = {H{\"a}gele, A. and Bakouch, E. and Kosson, A. and Ben Allal, L. and Von Werra, L. and Jaggi, M.},
  title         = {Scaling Laws and Compute-Optimal Training Beyond Fixed Training Durations},
  booktitle     = {Conference on Neural Information Processing Systems (NeurIPS)},
  year          = {2024}
}

@inproceedings{paszke2019pytorch,
  author        = {Paszke, A. and Gross, S. and Massa, F. and Lerer, A. and Bradbury, J. and Chanan, G. and Killeen, T. and Lin, Z. and Gimelshein, N. and Antiga, L. and Desmaison, A. and Kopf, A. and Yang, E. and DeVito, Z. and Raison, M. and Tejani, A. and Chilamkurthy, S. and Steiner, B. and Fang, L. and Bai, J. and Chintala, S.},
  title         = {{PyTorch}: An Imperative Style, High-Performance Deep Learning Library},
  booktitle     = {Conference on Neural Information Processing Systems (NeurIPS)},
  year          = {2019}
}

@inproceedings{wolf2020transformers,
  author        = {Wolf, T. and Debut, L. and Sanh, V. and Chaumond, J. and Delangue, C. and Moi, A. and Cistac, P. and Rault, T. and Louf, R. and Funtowicz, M. and Davison, J. and Shleifer, S. and von Platen, P. and Ma, C. and Jernite, Y. and Plu, J. and Xu, C. and Le Scao, T. and Gugger, S. and Drame, M. and Lhoest, Q. and Rush, A.},
  title         = {{Transformers}: State-of-the-Art Natural Language Processing},
  booktitle     = {Proceedings of the 2020 Conference on Empirical Methods in Natural Language Processing: System Demonstrations},
  year          = {2020}
}

@misc{chen2026reasoning,
  author        = {Chen, X. and Zhao, A. and Xia, H. and Lu, X. and Wang, H. and Chen, Y. and Zhang, W. and Wang, J. and Li, W. and Shen, X.},
  title         = {Reasoning Beyond Language: A Comprehensive Survey on Latent {Chain-of-Thought} Reasoning},
  eprint        = {2505.16782},
  archiveprefix = {arXiv},
  year          = {2025}
}

@misc{openai2024learning,
  title         = {Learning to reason with {LLMs}},
  author        = {{OpenAI}},
  url           = {https://openai.com/index/learning-to-reason-with-llms/},
  year          = {2024}
}

@misc{nye2021show,
  author        = {Nye, M. and Andreassen, A. J. and Gur-Ari, G. and Michalewski, H. and Austin, J. and Bieber, D. and Dohan, D. and Lewkowycz, A. and Bosma, M. and Luan, D. and Sutton, C. and Odena, A.},
  title         = {Show Your Work: Scratchpads for Intermediate Computation with Language Models},
  eprint        = {2112.00114},
  archiveprefix = {arXiv},
  year          = {2021}
}

@misc{deng2023implicit,
  author        = {Deng, Y. and Prasad, K. and Fernandez, R. and Smolensky, P. and Chaudhary, V. and Shieber, S.},
  title         = {Implicit chain of thought reasoning via knowledge distillation},
  eprint        = {2311.01460},
  archiveprefix = {arXiv},
  year          = {2023}
}

@misc{suleymanzade2026mux,
  author        = {Suleymanzade, A. and Gozeten, H. A. and Bronstein, M. and Ceylan, {\.I}. {\.I}. and Kim, J.},
  title         = {{MUX}: Continuous reasoning via multiplexed tokens},
  eprint        = {2607.18264},
  archiveprefix = {arXiv},
  year          = {2026}
}

@misc{fan2026bridging,
  author        = {Fan, Y. and Svete, A. and Lee, K.},
  title         = {Bridging the gap between latent and explicit reasoning with looped transformers},
  eprint        = {2606.31779},
  archiveprefix = {arXiv},
  year          = {2026}
}

@misc{tang2026multiplex,
  author        = {Tang, Y. and Dong, L. and Hao, Y. and Dong, Q. and Wei, F. and Gu, J.},
  title         = {Multiplex thinking: Reasoning via token-wise branch-and-merge},
  eprint        = {2601.08808},
  archiveprefix = {arXiv},
  year          = {2026}
}

@misc{wang2026gts,
  author        = {Wang, M. and Bai, Y. and Vu, T.-T. and Shareghi, E. and Haffari, G.},
  title         = {{GTS}: Inference-time scaling of latent reasoning with a learnable {Gaussian} thought sampler},
  eprint        = {2602.14077},
  archiveprefix = {arXiv},
  year          = {2026}
}

@misc{chen2021evaluating,
  author        = {Chen, M. and Tworek, J. and Jun, H. and Yuan, Q. and Pinto, H. P. de O. and Kaplan, J. and Edwards, H. and Burda, Y. and Joseph, N. and Brockman, G. and Ray, A. and Puri, R. and Krueger, G. and Petrov, M. and Khlaaf, H. and Sastry, G. and Mishkin, P. and Chan, B. and Gray, S. and Ryder, N. and Pavlov, M. and Power, A. and Kaiser, L. and Bavarian, M. and Winter, C. and Tillet, P. and Such, F. P. and Cummings, D. and Plappert, M. and Chantzis, F. and Barnes, E. and Herbert-Voss, A. and Guss, W. H. and Nichol, A. and Paino, A. and Tezak, N. and Tang, J. and Babuschkin, I. and Balaji, S. and Jain, S. and Saunders, W. and Hesse, C. and Carr, A. N. and Leike, J. and Achiam, J. and Misra, V. and Morikawa, E. and Radford, A. and Knight, M. and Brundage, M. and Murati, M. and Mayer, K. and Welinder, P. and McGrew, B. and Amodei, D. and McCandlish, S. and Sutskever, I. and Zaremba, W.},
  title         = {Evaluating Large Language Models Trained on Code},
  eprint        = {2107.03374},
  archiveprefix = {arXiv},
  year          = {2021}
}

@misc{li2025implicit,
  author        = {Li, J. and Fu, Y. and Fan, L. and Liu, J. and Shu, Y. and Qin, C. and Yang, M. and King, I. and Ying, R.},
  title         = {Implicit Reasoning in Large Language Models: A Comprehensive Survey},
  eprint        = {2509.02350},
  archiveprefix = {arXiv},
  year          = {2025}
}

@misc{radford2019language,
  author        = {Radford, A. and Wu, J. and Child, R. and Luan, D. and Amodei, D. and Sutskever, I.},
  title         = {Language Models are Unsupervised Multitask Learners},
  url           = {https://cdn.openai.com/better-language-models/language_models_are_unsupervised_multitask_learners.pdf},
  year          = {2019}
}

@misc{benallal2025smollm2,
  author        = {Ben Allal, L. and Lozhkov, A. and Bakouch, E. and Mart{\'i}n Bl{\'a}zquez, G. and Penedo, G. and Tunstall, L. and Marafioti, A. and Kydl{\'i}{\v c}ek, H. and Piqueres Lajar{\'i}n, A. and Srivastav, V. and Lochner, J. and Fahlgren, C. and Nguyen, X.-S. and Fourrier, C. and Burtenshaw, B. and Larcher, H. and Zhao, H. and Zakka, C. and Morlon, M. and Raffel, C. and von Werra, L. and Wolf, T.},
  title         = {{SmolLM2}: When Smol Goes Big -- Data-Centric Training of a Small Language Model},
  eprint        = {2502.02737},
  archiveprefix = {arXiv},
  year          = {2025}
}

@misc{lightman2023verify,
  author        = {Lightman, H. and Kosaraju, V. and Burda, Y. and Edwards, H. and Baker, B. and Lee, T. and Leike, J. and Schulman, J. and Sutskever, I. and Cobbe, K.},
  title         = {Let's Verify Step by Step},
  eprint        = {2305.20050},
  archiveprefix = {arXiv},
  year          = {2023}
}

@misc{tal2023scq5k,
  author        = {{TAL Education Group}},
  title         = {{TAL-SCQ5K}},
  url           = {https://huggingface.co/datasets/math-eval/TAL-SCQ5K},
  year          = {2023}
}

@misc{mario2024gaokao,
  author        = {{MARIO-Math-Reasoning}},
  title         = {{Gaokao2023-Math-En}},
  url           = {https://huggingface.co/datasets/MARIO-Math-Reasoning/Gaokao2023-Math-En},
  year          = {2024}
}

@misc{cobbe2021training,
  author        = {Cobbe, K. and Kosaraju, V. and Bavarian, M. and Chen, M. and Jun, H. and Kaiser, {\L}. and Plappert, M. and Tworek, J. and Hilton, J. and Nakano, R. and Hesse, C. and Schulman, J.},
  title         = {Training Verifiers to Solve Math Word Problems},
  eprint        = {2110.14168},
  archiveprefix = {arXiv},
  year          = {2021}
}

@misc{chen2025reasoning,
  author        = {Chen, Y. and Benton, J. and Radhakrishnan, A. and Uesato, J. and Denison, C. and Schulman, J. and Somani, A. and Hase, P. and Wagner, M. and Roger, F. and Mikulik, V. and Bowman, S. R. and Leike, J. and Kaplan, J. and Perez, E.},
  title         = {Reasoning Models Don't Always Say What They Think},
  eprint        = {2505.05410},
  archiveprefix = {arXiv},
  year          = {2025}
}

@misc{korbak2025chain,
  author        = {Korbak, T. and Balesni, M. and Barnes, E. and Bengio, Y. and Benton, J. and Bloom, J. and Chen, M. and Cooney, A. and Dafoe, A. and Dragan, A. and Emmons, S. and Evans, O. and Farhi, D. and Greenblatt, R. and Hendrycks, D. and Hobbhahn, M. and Hubinger, E. and Irving, G. and Jenner, E. and Kokotajlo, D. and Krakovna, V. and Legg, S. and Lindner, D. and Luan, D. and M{\k a}dry, A. and Michael, J. and Nanda, N. and Orr, D. and Pachocki, J. and Perez, E. and Phuong, M. and Roger, F. and Saxe, J. and Shlegeris, B. and Soto, M. and Steinberger, E. and Wang, J. and Zaremba, W. and Baker, B. and Shah, R. and Mikulik, V.},
  title         = {Chain of Thought Monitorability: A New and Fragile Opportunity for {AI} Safety},
  eprint        = {2507.11473},
  archiveprefix = {arXiv},
  year          = {2025}
}

@misc{hinton2015distilling,
  author        = {Hinton, G. and Vinyals, O. and Dean, J.},
  title         = {Distilling the Knowledge in a Neural Network},
  eprint        = {1503.02531},
  archiveprefix = {arXiv},
  year          = {2015}
}

@misc{lee2026flow,
  author        = {Lee, C. and Yoo, J. and Agarwal, M. and Shah, S. and Huang, J. and Raghunathan, A. and Hong, S. and Boffi, N. M. and Kim, J.},
  title         = {Flow Map Language Models: One-step Language Modeling via Continuous Denoising},
  eprint        = {2602.16813},
  archiveprefix = {arXiv},
  year          = {2026}
}

@misc{chen2026langflow,
  author        = {Chen, Y. and Liang, C. and Sui, H. and Guo, R. and Cheng, C. and You, J. and Liu, G.},
  title         = {{LangFlow}: Continuous Diffusion Rivals Discrete in Language Modeling},
  eprint        = {2604.11748},
  archiveprefix = {arXiv},
  year          = {2026}
}

@misc{jiang2026textldm,
  author        = {Jiang, J. and Ren, J. and Li, W. and Wang, B. and Sun, H. and Yang, Y. and Liu, J. and Zhang, Y. and Zheng, S. and Zhang, Y. and Huang, H. and Duan, N. and Zuo, W.},
  title         = {{TextLDM}: Language Modeling with Continuous Latent Diffusion},
  eprint        = {2605.07748},
  archiveprefix = {arXiv},
  year          = {2026}
}

@misc{hu2026elf,
  author        = {Hu, K. and Qiu, L. and Lu, Y. and Zhao, H. and Li, T. and Kim, Y. and Andreas, J. and He, K.},
  title         = {{ELF}: Embedded Language Flows},
  eprint        = {2605.10938},
  archiveprefix = {arXiv},
  year          = {2026}
}

@misc{xu2025softcotpp,
  author        = {Xu, Y. and Guo, X. and Zeng, Z. and Miao, C.},
  title         = {{SoftCoT++}: Test-Time Scaling with Soft {Chain-of-Thought} Reasoning},
  eprint        = {2505.11484},
  archiveprefix = {arXiv},
  year          = {2025}
}

@misc{chen2025think,
  author        = {Chen, Z. and Cui, S. and Ye, D. and Zhang, Y. and Bian, Y. and Zhu, T.},
  title         = {Think Consistently, Reason Efficiently: Energy-Based Calibration for Implicit {Chain-of-Thought}},
  eprint        = {2511.07124},
  archiveprefix = {arXiv},
  year          = {2025}
}

@misc{zhang2025days,
  author        = {Zhang, C. and Deng, Y. and Lin, X. and Wang, B. and Ng, D. and Ye, H. and Li, X. and Xiao, Y. and Mo, Z. and Zhang, Q. and Bing, L.},
  title         = {100 Days After {DeepSeek-R1}: A Survey on Replication Studies and More Directions for Reasoning Language Models},
  eprint        = {2505.00551},
  archiveprefix = {arXiv},
  year          = {2025}
}

@misc{kydlicek2025mathverify,
  author        = {Kydl{\'i}{\v c}ek, H.},
  title         = {{Math-Verify}: Math Verification Library},
  url           = {https://github.com/huggingface/Math-Verify/releases/tag/0.8.0},
  year          = {2025},
  note          = {Version 0.8.0}
}

@misc{gao2022pal,
  author        = {Gao, L. and Madaan, A. and Zhou, S. and Alon, U. and Liu, P. and Yang, Y. and Callan, J. and Neubig, G.},
  title         = {{PAL}: Program-aided Language Models},
  eprint        = {2211.10435},
  archiveprefix = {arXiv},
  year          = {2022}
}

@misc{hu2024minicpm,
  author        = {Hu, S. and Tu, Y. and Han, X. and He, C. and Cui, G. and Long, X. and Zheng, Z. and Fang, Y. and Huang, Y. and Zhao, W. and Zhang, X. and Thai, Z. L. and Zhang, K. and Wang, C. and Yao, Y. and Zhao, C. and Zhou, J. and Cai, J. and Zhai, Z. and Ding, N. and Jia, C. and Zeng, G. and Li, D. and Liu, Z. and Sun, M.},
  title         = {{MiniCPM}: Unveiling the Potential of Small Language Models with Scalable Training Strategies},
  eprint        = {2404.06395},
  archiveprefix = {arXiv},
  year          = {2024}
}

@misc{mangrulkar2022peft,
  author        = {Mangrulkar, S. and Gugger, S. and Debut, L. and Belkada, Y. and Paul, S. and Bossan, B. and Tietz, M.},
  title         = {{PEFT}: State-of-the-art Parameter-Efficient Fine-Tuning methods},
  url           = {https://github.com/huggingface/peft},
  year          = {2022}
}

\clearpage
\appendix
\section{Additional Related Work}\label{sec:additional-related-work}

\paragraph{Distribution distillation}
Since \emph{knowledge distillation} for neural networks gained popularity~\citep{hinton2015distilling}, subsequent studies explored \emph{distribution distillation} as well, modeling teacher prediction distributions rather than their means~\citep{malinin2020ensemble,cui2022accelerating}.
More recently, flow matching has been used to model ensemble logit distributions~\citep{park2025ensemble}.
They share our goal of distilling an empirical distribution, but concern predictive outputs rather than generative language modeling.
In non-latent language models, multiple teacher-sampled textual rationales per query were distilled~\citep{li2023symbolic,ho2023large}, with consistency supervision across rationale-conditioned answer distributions~\citep{chen2023mcc}.
In latent reasoning, prior work noted that mean-squared matching can average teacher states across paths without yielding a valid path, and used layerwise mixtures instead~\citep{deng2023implicit}.
However, to the best of our knowledge, no prior work explicitly distills per-query empirical distributions of complete textual CoT rationales into conditional continuous rationale distributions.

\paragraph{Flow matching for language modeling}
Flow matching has been applied to text generation in continuous embedding space, including question generation, paraphrasing, and dialogue~\citep{hu2024flow}, and non-autoregressive language modeling directly in discrete token space~\citep{gat2024discrete}.
Recent studies learn continuous flows over one-hot token representations and distill these flows for one-step generation~\citep{lee2026flow}, connect embedding-space language models to flow matching~\citep{chen2026langflow}, and perform flow matching in continuous latents learned by text autoencoders~\citep{jiang2026textldm}.
Another study keeps generation in continuous embedding space until the final step, then maps the embeddings to discrete tokens using a shared network for denoising and token prediction~\citep{hu2026elf}.
These approaches ultimately aim to generate verbalized text; even their continuous representations are intended to be decoded into tokens.
On the other hand, in our approach, sampled rationales remain continuous and serve as contexts for answer generation without first decoding these latent states into text.

\paragraph{Latent augmentation of textual reasoning}
A complementary line of work uses continuous latents to assist textual reasoning rather than fully replace it.
One study generates soft thoughts with a frozen assistant and projects them into a frozen LLM's input space before textual CoT generation~\citep{xu2025softcot}.
The latent prefix thus conditions the decoder, which still produces an explicit rationale and answer.
Subsequent work diversifies these prefixes with specialized initial tokens and contrastive learning~\citep{xu2025softcotpp}, or refines them with an energy model before textual reasoning~\citep{chen2025think}.
Another study augments a frozen decoder's key-value cache with latent embeddings from a coprocessor trained on standard pretraining data with the decoder's language modeling loss, improving subsequent decoding without task-specific training~\citep{liu2025deliberation}.
These methods use latents to condition textual generation.
We instead distill complete rationale outcomes into continuous rationale distributions and decode answers directly from sampled contexts without first generating a textual reasoning trace.

\newcounter{appendixdefinition}[section]
\renewcommand{\theappendixdefinition}{\thesection.\arabic{appendixdefinition}}
\newcounter{appendixproposition}[section]
\renewcommand{\theappendixproposition}{\thesection.\arabic{appendixproposition}}
\newcounter{appendixcorollary}[section]
\renewcommand{\theappendixcorollary}{\thesection.\arabic{appendixcorollary}}

\section{Analysis of Within-Query Mixup}
\label{sec:theoretical-analysis}

We analyze how within-query mixup changes the reference targets and the supervision used to learn from them.
Theoretically, it contracts target covariance while retaining supervision on the original targets, and induces an exact interpolation of linear-path flow-matching regression examples.
We then distinguish this reference geometry from the rationales generated under joint training of the sampler and contextualizer.
Section~\ref{sec:mixup-empirical-evidence} relates the analysis to the trajectory-level probe in Section~\ref{par:trajectory-level-mixup-compatibility}, the mixing-weight ablation in Section~\ref{par:mixing-weight-distribution}, and generated-latent dispersion measurements.
These observations do not establish a causal link between latent variance and answer coverage.
Throughout this section, we fix $q$ and suppress it in the notation.

\subsection{Reference geometry}
\label{sec:mixup-reference-geometry}

\refstepcounter{appendixdefinition}\label{def:reference-distributions}
\paragraph{Definition \theappendixdefinition\ (Reference distributions).}
Let $\{\bm z^{(m)}\in\mathbb R^{S\times D}\}_{m=1}^{M}$ be the teacher rationale targets obtained by position-weighted embedding pooling, with $M \ge 2$.
The Gaussian source and empirical target distributions are given by:
\begin{equation}\textstyle
\textstyle P_0 =\mathcal N(\bm0,I),
\quad
\textstyle \widehat P_1 =\frac1M\sum_{m=1}^M\delta_{\bm z^{(m)}}.
\end{equation}
Draw $(j,k)$ uniformly from the ordered pairs with $j\neq k$, and independently draw $\lambda\sim\nu$, where $\nu$ is supported on $[0,1]$.
Within-query mixup yields:
\begin{equation}\textstyle
\begin{aligned}
\textstyle \bm z_{\mathrm{mix}}&\textstyle =\lambda\bm z^{(j)}+(1-\lambda)\bm z^{(k)},\\
\textstyle P_{\mathrm{mix}}&\textstyle =\operatorname{Law}(\bm z_{\mathrm{mix}}),
\end{aligned}
\label{eq:mixup-reference}
\end{equation}
and introduces the following target distribution:
\begin{equation}\textstyle
\widetilde P_1\textstyle =(1-\rho)\widehat P_1+\rho P_{\mathrm{mix}},
\end{equation}
where $\rho\in[0,1]$ is the probability of selecting a mixed instance (throughout our experiments, $\rho=1/2$).
Training on both original and mixed targets corresponds to $0<\rho<1$.
No mixing is recovered with $\rho=0$, or equivalently with $\lambda=1$ almost surely.
The flow target is sampled as $\bm z_1\sim P_1$, where $P_1=\widehat P_1$ for training without mixup and $P_1=\widetilde P_1$ for augmented training.
The source $\bm z_0\sim P_0$ is independent of the target sampling.

\refstepcounter{appendixproposition}\label{prop:mixup-reference-moments}
\paragraph{Proposition \theappendixproposition\ (Mean preservation and covariance contraction).}
Under Definition~\ref{def:reference-distributions}, let $\bm\mu$ and $\Sigma$ denote the mean and covariance of the vectorized teacher targets under $\widehat P_1$, and define $\beta_\nu=\mathbb E_\nu[\lambda(1-\lambda)]$.
Both $P_{\mathrm{mix}}$ and $\widetilde P_1$ preserve $\bm\mu$ and have covariances:
\begin{equation}\textstyle
\begin{aligned}
\textstyle \operatorname{Cov}(P_{\mathrm{mix}})
&\textstyle =\left(1-\frac{2M\beta_\nu}{M-1}\right)\Sigma,\\
\textstyle \operatorname{Cov}(\widetilde P_1)
&\textstyle =\left(1-\rho\frac{2M\beta_\nu}{M-1}\right)\Sigma.
\end{aligned}
\label{eq:mixup-covariance}
\end{equation}
All covariances are computed after vectorization.
Since $0\leq\beta_\nu\leq1/4$ and $M\geq2$, both covariance factors lie in $[0,1]$.

\emph{Proof.}
The indices $j$ and $k$ each have a uniform marginal and are independent of $\lambda$.
Their weighted combination therefore has mean $\bm\mu$.
Set $\bm c^{(m)}=\operatorname{vec}(\bm z^{(m)})-\bm\mu$.
Uniform sampling of distinct pairs and $\sum_m\bm c^{(m)}=0$ give
\begin{equation}\textstyle
\mathbb E[\bm c^{(j)}\bm c^{(k)\top}]
=\frac{\sum_{j\neq k}\bm c^{(j)}\bm c^{(k)\top}}{M(M-1)}
=-\frac{\Sigma}{M-1}.
\end{equation}
Expanding the covariance of $\lambda\bm c^{(j)}+(1-\lambda)\bm c^{(k)}$ gives $(1-2\beta_\nu)\Sigma-2\beta_\nu\Sigma/(M-1)$, which yields the first identity.
The original and mixed distributions share mean $\bm\mu$, so their mixture has covariance $(1-\rho)\Sigma+\rho\operatorname{Cov}(P_{\mathrm{mix}})$, as claimed.\hfill$\square$

\emph{Remark.}
Augmentation preserves the target mean.
For $\rho\beta_\nu>0$, it strictly reduces variance in every direction with positive original variance, consistent with prior analyses~\citep{lee2025generalized}.

\refstepcounter{appendixcorollary}\label{cor:retained-supervision}
\paragraph{Corollary \theappendixcorollary\ (Retention of original supervision).}
Under Definition~\ref{def:reference-distributions}, for $0\leq\rho<1$,
\begin{equation}\textstyle
(1-\rho)\Sigma\preceq\operatorname{Cov}(\widetilde P_1)\preceq\Sigma,
\label{eq:retained-covariance}
\end{equation}
where $\preceq$ denotes the positive semidefinite order.
The covariance rank is unchanged, and for any measurable region $B$ in rationale space,
\begin{equation}\textstyle
\widetilde P_1(B)\geq(1-\rho)\widehat P_1(B).
\label{eq:retained-path-mass}
\end{equation}

\emph{Proof.}
Proposition~\ref{prop:mixup-reference-moments} gives $0\preceq\operatorname{Cov}(P_{\mathrm{mix}})\preceq\Sigma$.
The two target distributions share their mean, so their mixture satisfies Equation~\ref{eq:retained-covariance}.
Since $1-\rho>0$, its covariance has the same null space as $\Sigma$.
Equation~\ref{eq:retained-path-mass} follows from the mixture definition and $P_{\mathrm{mix}}(B)\geq0$.\hfill$\square$

\emph{Remark.}
Each original teacher component contributes weight $(1-\rho)/M$ to the augmented reference.
The reference therefore retains supervision on every observed target and at least a fraction $1-\rho$ of the original variance in every direction.
The augmentation can thus add targets between observed rationales while reducing their covariance; all targets remain within the original convex hull.

\paragraph{Mixing weights.}
For $\nu=\operatorname{Beta}(\alpha,\alpha)$, the mixing-weight variance is $\operatorname{Var}(\lambda)=1/[4(2\alpha+1)]$, so
\begin{equation}\textstyle
\beta_\nu=\frac14-\operatorname{Var}(\lambda)
=\frac{\alpha}{2(2\alpha+1)}.
\end{equation}
The choices $\alpha=2$, $\alpha=5$, and deterministic midpoint mixing ($\alpha\to\infty$) therefore give $\beta_\nu=1/5$, $5/22$, and $1/4$, respectively.
No mixing, represented by $\lambda=1$ almost surely, gives $\beta_\nu=0$.
For fixed teacher targets and $\rho>0$, the covariance factor of $\widetilde P_1$ in Proposition~\ref{prop:mixup-reference-moments} therefore decreases in the order no mixing, $\operatorname{Beta}(2,2)$, $\operatorname{Beta}(5,5)$, and midpoint mixing.
Both this ordering and the retention guarantees characterize reference supervision; they do not determine generated-latent dispersion or answer coverage under joint training.

\subsection{Effect on the training objective}
\label{sec:mixup-training-objective}

For linear-path flow matching, target mixup induces a corresponding interpolation of regression inputs and velocity targets.

\refstepcounter{appendixproposition}\label{prop:mixup-fm-supervision}
\paragraph{Proposition \theappendixproposition\ (Interpolation of flow matching supervision).}
Fix a query, a time $t$, and a common source sample $\bm z_0$.
For $m\in\{j,k\}$, define the original training states and velocity targets by
\begin{equation}\textstyle
\begin{aligned}
\textstyle \bm z_t^{(m)}&\textstyle =(1-t)\bm z_0+t\bm z^{(m)},\\
\textstyle \bm d^{(m)}&\textstyle =\bm z^{(m)}-\bm z_0.
\end{aligned}
\end{equation}
The mixed target $\bm z_{\mathrm{mix}}$ induces $\bm z_t^{\mathrm{mix}}=(1-t)\bm z_0+t\bm z_{\mathrm{mix}}$ and $\bm d^{\mathrm{mix}}=\bm z_{\mathrm{mix}}-\bm z_0$, which satisfy
\begin{equation}\textstyle
\begin{aligned}
\textstyle \bm z_t^{\mathrm{mix}}&\textstyle =\lambda\bm z_t^{(j)}+(1-\lambda)\bm z_t^{(k)},\\
\textstyle \bm d^{\mathrm{mix}}&\textstyle =\lambda\bm d^{(j)}+(1-\lambda)\bm d^{(k)}.
\end{aligned}
\label{eq:fm-mixup-identity}
\end{equation}
\emph{Proof.}
Substitute $\bm z_{\mathrm{mix}}=\lambda\bm z^{(j)}+(1-\lambda)\bm z^{(k)}$ into the mixed state and velocity target, and use $\lambda+(1-\lambda)=1$ in both expressions.\hfill$\square$

\emph{Remark.}
The shared source sample is an analytical coupling; training can draw fresh noise independently for every example.
Proposition~\ref{prop:mixup-fm-supervision} shows that target augmentation interpolates both sides of the FM regression problem while keeping the query and time fixed.

\paragraph{Regularization interpretation.}
Proposition~\ref{prop:mixup-fm-supervision} connects the FM construction to supervised mixup~\citep{zhang2018mixup,carratino2022mixup}.
With finite teacher samples and limited model capacity, supplying regression examples between observed targets may discourage over-specialization to their embeddings.
This interpretation concerns augmentation of FM supervision and does not require the learned velocity field to be linear.
The mixing-weight ablation in Section~\ref{par:mixing-weight-distribution} provides empirical support for this regularization interpretation.

\paragraph{Role of auxiliary supervision.}
The answer and local content losses in Section~\ref{sec:answer-aware-contextualization} supervise generated rationales $\hat{\bm z}_1=F_\theta(\bm z_0,q)$.
At fixed parameters, these samples depend on the source noise and query, and are independent of the sampled teacher indices and mixing weight.
Let $\ell_m(\bm z_0)$ denote either auxiliary loss evaluated on the contextualized generated rationale using supervision from teacher trace $m$.
For fixed $\bm z_0$, uniform index marginals and independence of $\lambda$ give
\begin{equation}\textstyle
\begin{aligned}
\textstyle \mathbb E_{j,k,\lambda}\big[\lambda\ell_j(\bm z_0)
&\textstyle +(1-\lambda)\ell_k(\bm z_0)\big]\\
&\textstyle =\frac1M\sum_{m=1}^M\ell_m(\bm z_0).
\end{aligned}
\label{eq:mixed-auxiliary-expectation}
\end{equation}
For the same teacher set and query distribution, mixup therefore changes the expected FM objective while preserving the expected auxiliary objective at fixed parameters.

\paragraph{Reference targets and generated rationales.}
The sampler is optimized under the joint objective in Equation~\ref{eq:joint-objective}, with both auxiliary losses backpropagating through the contextualizer and ODE solver.
Its generated rationale distribution therefore need not match the reference target distribution or inherit its covariance ordering, even though the expected auxiliary objective is unchanged by mixing at fixed parameters.
Finite model capacity, optimization error, and numerical integration can also affect this correspondence, including under flow-only training.
Section~\ref{sec:mixup-empirical-evidence} examines generated-latent dispersion and answer coverage separately to relate this distinction to the learned reasoner's behavior.

\subsection{Empirical evidence}
\label{sec:mixup-empirical-evidence}

The preceding analysis characterizes target geometry and FM supervision while distinguishing them from the behavior of the jointly trained reasoner.
We examine the empirical implications through answer coverage, generated-latent dispersion, and a controlled probe of trajectory-level interpolation.

\paragraph{Mixing-weight distribution.}
Among the mixing-weight settings in Table~\ref{tab:ablation}, $\operatorname{Beta}(2,2)$ achieves the highest arithmetic \texttt{pass@512}, whereas $\operatorname{Beta}(5,5)$ achieves the highest advanced-mathematics coverage.
Midpoint mixing ($\lambda=1/2$) improves advanced coverage over no mixing but does not improve arithmetic coverage.
Coverage therefore does not improve monotonically with stronger reference-target covariance contraction in Proposition~\ref{prop:mixup-reference-moments}.
Nevertheless, both stochastic mixing settings consistently improve coverage over no mixing across the two benchmark groups, lending support to the regularization interpretation in Section~\ref{sec:mixup-training-objective}.

\paragraph{Generated-latent dispersion.}
We next compare the dispersion of generated flow endpoints $\hat{\bm z}_1$ under no mixing and midpoint mixing ($\lambda=1/2$), which maximizes reference-target covariance contraction at fixed $\rho$ (Proposition~\ref{prop:mixup-reference-moments}).
For each of 500 randomly subsampled \texttt{GSM8K} test queries, we draw 128 Gaussian initial states and use the same set for both models to generate flow endpoints.
For each query, we compute the sample variance across the 128 endpoints at each latent position and feature dimension, then average over all positions and dimensions.
Averaging these query-level variances yields $1.7259\times10^{-2}$ for no mixing and $1.7257\times10^{-2}$ for midpoint mixing.
Thus, even the strongest reference-target contraction is accompanied by nearly unchanged mean endpoint variance in this comparison.
This observation is consistent with Section~\ref{sec:mixup-training-objective}: flow matching and auxiliary supervision jointly shape the sampler, so target covariance alone does not determine the dispersion of its generated rationales.
These measurements characterize generated-latent dispersion; the mixing-weight ablation in Table~\ref{tab:ablation} provides separate empirical evidence that mixing can improve answer coverage.

\begin{table*}[t]
\centering
\setlength{\aboverulesep}{0pt}
\setlength{\belowrulesep}{0pt}
\setlength{\tabcolsep}{0.5em}
\renewcommand{\arraystretch}{1.1}
\resizebox{\textwidth}{!}{\begin{tabular}{llll}
  \toprule
  Resource & Reference & Hugging Face repository & Reported license \\
  \midrule
  \rowcolor{black!4}\multicolumn{4}{c}{\textbf{\textit{Model backbones}}} \\
  \texttt{GPT-2-0.1B} & \citet{radford2019language}
    & \href{https://huggingface.co/openai-community/gpt2}{\nolinkurl{openai-community/gpt2}}
    & MIT \\
  \texttt{SLM-2-1.7B} & \citet{benallal2025smollm2}
    & \href{https://huggingface.co/HuggingFaceTB/SmolLM2-1.7B}{\nolinkurl{HuggingFaceTB/SmolLM2-1.7B}}
    & Apache-2.0 \\
  \midrule
  \rowcolor{black!4}\multicolumn{4}{c}{\textbf{\textit{Training data sources}}} \\
  \texttt{GSM8K-Aug} & \citet{deng2023implicit,wu2025parallel}
    & \href{https://huggingface.co/datasets/whynlp/gsm8k-aug}{\nolinkurl{whynlp/gsm8k-aug}}
    & Not stated$^{\dagger}$ \\
  \texttt{GSM8K-Aug-NL} & \citet{deng2023implicit,wu2025parallel}
    & \href{https://huggingface.co/datasets/whynlp/gsm8k-aug-nl}{\nolinkurl{whynlp/gsm8k-aug-nl}}
    & Not stated$^{\dagger}$ \\
  \midrule
  \rowcolor{black!4}\multicolumn{4}{c}{\textbf{\textit{Evaluation benchmarks}}} \\
  \texttt{GSM8K} & \citet{cobbe2021training}
    & \href{https://huggingface.co/datasets/openai/gsm8k}{\nolinkurl{openai/gsm8k}}
    & MIT \\
  \texttt{SVAMP} & \citet{patel2021nlp}
    & \href{https://huggingface.co/datasets/ChilleD/SVAMP}{\nolinkurl{ChilleD/SVAMP}}
    & MIT \\
  \texttt{GSMHard} & \citet{gao2022pal}
    & \href{https://huggingface.co/datasets/reasoning-machines/gsm-hard}{\nolinkurl{reasoning-machines/gsm-hard}}
    & MIT \\
  \texttt{TAL} & \citet{tal2023scq5k}
    & \href{https://huggingface.co/datasets/math-eval/TAL-SCQ5K}{\nolinkurl{math-eval/TAL-SCQ5K}}
    & MIT \\
  \texttt{MATH500} & \citet{hendrycks2021measuring,lightman2023verify}
    & \href{https://huggingface.co/datasets/math-ai/math500}{\nolinkurl{math-ai/math500}}
    & Not stated$^{\S}$ \\
  \texttt{Gaokao23} & \citet{mario2024gaokao}
    & \href{https://huggingface.co/datasets/MARIO-Math-Reasoning/Gaokao2023-Math-En}{\nolinkurl{MARIO-Math-Reasoning/Gaokao2023-Math-En}}
    & MIT \\
  \bottomrule
\end{tabular}}
\par\smallskip
\begin{minipage}{0.96\textwidth}
\footnotesize
$^{\dagger}$ Hugging Face dataset card does not declare a license; associated \href{https://github.com/whynlp/pccot/blob/dev-publish/LICENSE}{PCCoT} repository carries an MIT license.
\par
$^{\S}$ Hugging Face dataset card does not declare a license; associated \href{https://github.com/hendrycks/math/blob/master/LICENSE}{\texttt{MATH}} and \href{https://github.com/openai/prm800k/blob/main/LICENSE}{\texttt{PRM800K}} repositories carry MIT licenses.
\end{minipage}
\caption{\textbf{Model backbones and dataset sources.} References, public Hugging Face releases, and reported licenses are listed for each resource. Licenses are those declared by the linked release; ``not stated'' means no license declaration was found in its card or repository metadata.}
\label{tab:resources}
\end{table*}

\paragraph{Trajectory-level compatibility.}
We also consider the controlled \texttt{CoT} probe in Section~\ref{par:trajectory-level-mixup-compatibility}, which tests answer compatibility when interpolating uncompressed rationale embeddings.
The probe uses 10,000 triples $(q,r_1,r_2)$ with two answer-correct rationales that have matching numbers of reasoning steps and token counts at each step.
Accuracy is 100\% with the original rationale $r_1$, 99.99\% with within-query midpoint interpolation, and 44.33\% when $r_2$ is replaced by a rationale from another query.
This contrast supports the choice to interpolate rationales belonging to the same query.
The compatibility evidence concerns aligned, uncompressed representations, while the mixing-weight ablation assesses the benefit of target interpolation in the jointly trained \texttt{RDD} model.

\section{Implementation Details}
\label{sec:implementation-details}

\subsection{Models and datasets}
\label{sec:models-and-datasets}

Table~\ref{tab:resources} lists the model and dataset sources used in our experiments, with bibliographic references, Hugging Face links, and reported licenses.

\paragraph{Training data preparation}
We sanitize and decontaminate the \texttt{GSM8K-Aug} and \texttt{GSM8K-Aug-NL} datasets;
the resulting data comprise 378,647 training instances and 500 validation instances.
To build data with multiple CoT reasoning paths per question, we sample 10 responses per training question from the textual \texttt{CoT} model of each backbone with temperature 1.0, top-$k$ 50, top-$p$ 1.0, and min-$p$ 0.1.
Any response that is truncated before a stop token or violates the response format is resampled, and questions still lacking 10 distinct responses after six attempts are resampled with temperature 2.0 and no min-$p$.
We then keep only the CoT reasoning paths whose final answer is correct.

\paragraph{Evaluation data preparation}
\texttt{TAL-SCQ5K} provides \texttt{EN} and \texttt{CN} subsets; we use only \texttt{EN} instances.
\texttt{Gaokao2023-Math-En} also contains AMC23 and ACT23 problems; we filter out those instances.
We consolidate all the listed benchmark datasets into a common QA format and apply answer normalization for verification with \texttt{math-verify}~\citep[v0.8.0; Apache-2.0;][]{kydlicek2025mathverify}.

\paragraph{Textual CoT model preparation}
For each backbone, all methods in our experiments are initialized from a textual \texttt{CoT} model, which also serves as the frozen ALM teacher for generating multiple textual CoT paths per query.
We add the special tokens \texttt{<|beginofthink|>} and \texttt{<|endofthink|>} and format each (question, rationale, answer) training instance as follows:
\begin{quote}
\small\raggedright
\microtypesetup{protrusion=false}
\verb!{question}\n!\allowbreak%
\verb!<|beginofthink|>\n!\allowbreak%
\verb!{rationale}\n!\allowbreak%
\verb!<|endofthink|>\n\n!\allowbreak%
\verb!{answer}<|endoftext|>!
\end{quote}
Here, \texttt{\{rationale\}} joins reasoning steps with two newline characters (\verb|\n\n|).

\subsection{Hyperparameters}
\label{sec:hyperparameters}

\begin{table}[t]
\centering
\setlength{\aboverulesep}{0pt}
\setlength{\belowrulesep}{0pt}
\setlength{\tabcolsep}{0.5em}
\renewcommand{\arraystretch}{1.1}
\resizebox{\linewidth}{!}{\begin{tabular}{l@{\hspace{4em}}c}
  \toprule
  Hyperparameter & Value \\
  \midrule
  \rowcolor{black!4}\multicolumn{2}{c}{\textbf{\textit{Optimization}}} \\
  Global batch size & 128 \\
  Optimizer & AdamW \\
  $(\beta_1, \beta_2)$ & $(0.9, 0.95)$ \\
  Weight decay & 0.1 \\
  Peak learning rate & $10^{-3}$ ($A$),\, $4{\times}10^{-3}$ ($B$) \\
  Schedule & WSD \\
  Warmup\,/\,decay & 40 steps\,/\,last 20\% \\
  Gradient clipping & 1.0 \\
  Precision & bf16 \\
  \midrule
  \rowcolor{black!4}\multicolumn{2}{c}{\textbf{\textit{LoRA}}} \\
  Rank & 128 \\
  $\alpha$ & 32 \\
  Dropout & 0.1 \\
  Target modules & all linear layers \\
  \bottomrule
\end{tabular}}
\caption{\textbf{Training hyperparameters.} Values shared across methods and backbones.}
\label{tab:hyperparameters}
\end{table}

\paragraph{Optimization}
We apply LoRA~\citep{hu2022lora} to all linear layers for both textual \texttt{CoT} training and latent reasoning fine-tuning.
Following LoRA+~\citep{hayou2024loraplus}, we set the learning rate for the $B$ matrices to four times that for the $A$ matrices.
We optimize the trainable parameters with AdamW~\citep{loshchilov2019decoupled} using a warmup--stable--decay~\citep[WSD;][]{hu2024minicpm} schedule with linear warmup and a $1-\sqrt{\cdot}$ decay to zero~\citep{hagele2024scaling}.
Table~\ref{tab:hyperparameters} lists the optimization hyperparameters shared across methods.

\paragraph{Textual \texttt{CoT}}
We train the textual \texttt{CoT} models for 300k steps with \texttt{GPT-2-0.1B} and 50k steps with \texttt{SLM-2-1.7B}.
In addition to the LoRA parameters, we train the embedding weights of the \texttt{<|beginofthink|>} and \texttt{<|endofthink|>} tokens.

\paragraph{Latent reasoning methods}
Starting from the textual \texttt{CoT} model of each backbone, we fine-tune all latent reasoning methods for 100k steps with \texttt{GPT-2-0.1B} and 10k steps with \texttt{SLM-2-1.7B}.
For \texttt{CODI}, \texttt{PCCoT}, and \texttt{LOTUS}, we follow the optimization settings in their original papers; \texttt{LOTUS} uses full fine-tuning instead of LoRA fine-tuning.
For \texttt{PCCoT} and \texttt{LOTUS}, we introduce the special token \texttt{<|think|>}, which corresponds to \texttt{<latent>} and \texttt{<lat>} in \citet{wu2025parallel} and \citet{fan2026bridging}, respectively.
In \texttt{PCCoT}, the embedding weights of this token are trained jointly with the LoRA parameters, while all other parameters remain frozen.

\paragraph{RDD}
The ODE solver integrates the velocity field with six Euler steps (NFE=6) during both training and inference.
For within-query mixup, $\lambda\sim\operatorname{Beta}(2,2)$ is drawn independently for each mixed target ($\operatorname{Beta}(0.5,0.5)$ for \texttt{SLM-2-1.7B}).
Each teacher rationale is split into $S=6$ chunks of contiguous reasoning steps with nearly equal step counts.
Within a chunk of $n_l$ tokens, the token at position $i\in\{0,\dots,n_l-1\}$ receives a weight proportional to $\exp\!\big(\sin(\tfrac{\pi i}{2(n_l-1)})\big)$, emphasizing later tokens; the normalized weights pool the chunk into its rationale vector and define its local token distribution $P_l$.
The reasoning adapter $\phi$ is active only in the sampler and contextualizer; the answer decoder uses the frozen ALM with the adapter disabled.
We use $\lambda_{\text{flow}}=1$ and $\lambda_{\text{loc}}=1$ throughout experiments.

\subsection{Evaluation}
\label{sec:evaluation}

\paragraph{\texttt{pass@}$k$}
For each query $q$, let $c_q$ be the number of correct answers among $n=512$ rollouts sampled at temperature 0.7.
We use the unbiased estimator~\citep{chen2021evaluating}, for $1\le k\le n$,
\begin{equation}\textstyle
\texttt{pass@}k(q)=1-\binom{n-c_q}{k}\big/\binom{n}{k}.
\label{eq:pass-at-k}
\end{equation}

\paragraph{Fidelity metrics}
For each query $q$, we construct empirical answer distributions $P_q$ and $Q_q$ for textual \texttt{CoT} and the evaluated method, respectively, using 512 rollouts from each.
The distributions are normalized frequencies over extracted answer categories; equivalent numeric answers share a category, while unextractable answers form a shared $\bot$ category.
Incorrect answers are retained, and correctness labels are not used to construct these distributions.
Let $M_q=(P_q+Q_q)/2$.
We compute Jensen--Shannon divergence~\citep[JSD;][]{lin1991divergence} and total variation~\citep[TV;][]{gibbs2002choosing} as
\begin{equation}\textstyle
\operatorname{JSD}(P_q,Q_q)=\frac{D_{\mathrm{KL}}(P_q\Vert M_q)+D_{\mathrm{KL}}(Q_q\Vert M_q)}{2},
\label{eq:answer-jsd}
\end{equation}
\begin{equation}\textstyle
\operatorname{TV}(P_q,Q_q)=\frac12\sum_a\lvert P_q(a)-Q_q(a)\rvert.
\label{eq:answer-tv}
\end{equation}
We use unsmoothed frequencies, base-2 logarithms in $D_{\mathrm{KL}}$, and the convention $0\log 0=0$.
Teacher support coverage (TSC) is the empirical teacher probability mass assigned to answers observed at least once from the evaluated method:
\begin{equation}\textstyle
\operatorname{TSC}(P_q,Q_q)=\sum_{a:\,Q_q(a)>0}P_q(a).
\label{eq:teacher-support-coverage}
\end{equation}
All three metrics lie in $[0,1]$.
Lower JSD and TV indicate closer answer distributions, while higher TSC indicates less missing teacher mass.

\subsection{Software and hardware}
\label{sec:compute}

The main libraries used in our experiments are \texttt{torch}~\citep[v2.11.0; BSD-3-Clause;][]{paszke2019pytorch}, \texttt{transformers}~\citep[v5.14.1; Apache-2.0;][]{wolf2020transformers}, and \texttt{peft}~\citep[v0.20.0; Apache-2.0;][]{mangrulkar2022peft}.
All experiments were run on NVIDIA RTX 3090 and RTX A6000 GPUs.

\section{Supplementary Results}
\label{sec:supplementary-results}

\begin{figure*}[t]
  \centering
  \includegraphics[width=\textwidth]{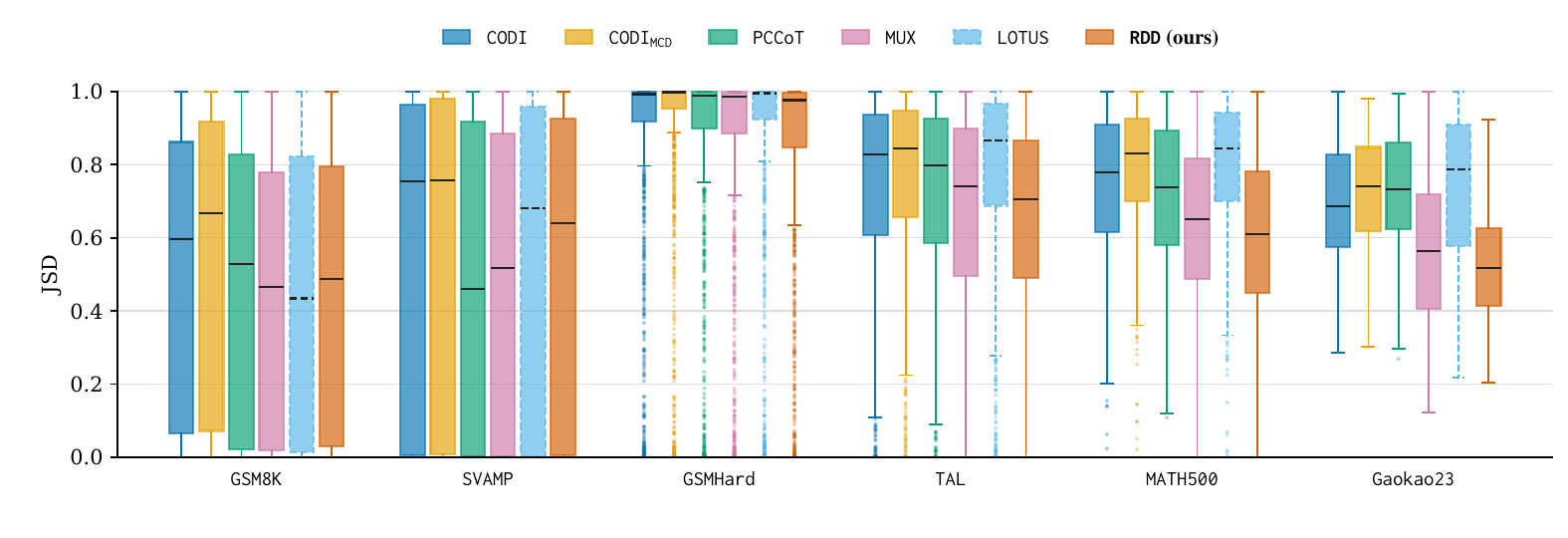}
  \par\vspace{0.4em}
  \includegraphics[width=\textwidth]{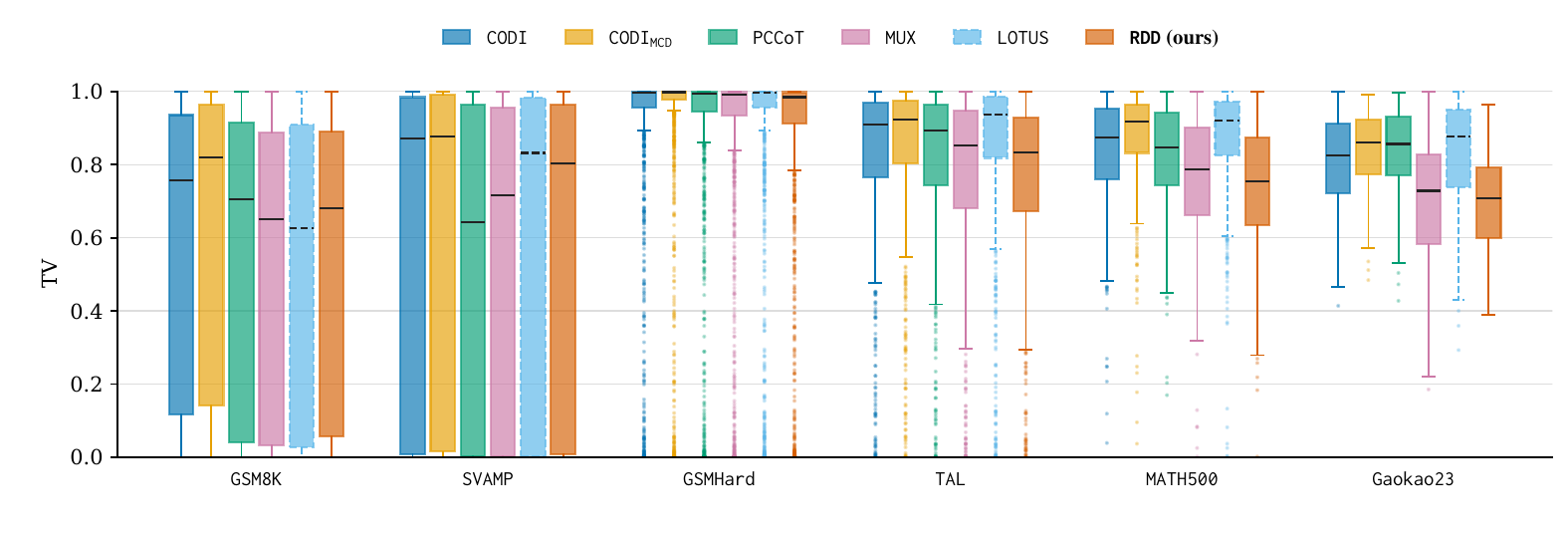}
  \par\vspace{0.4em}
  \includegraphics[width=\textwidth]{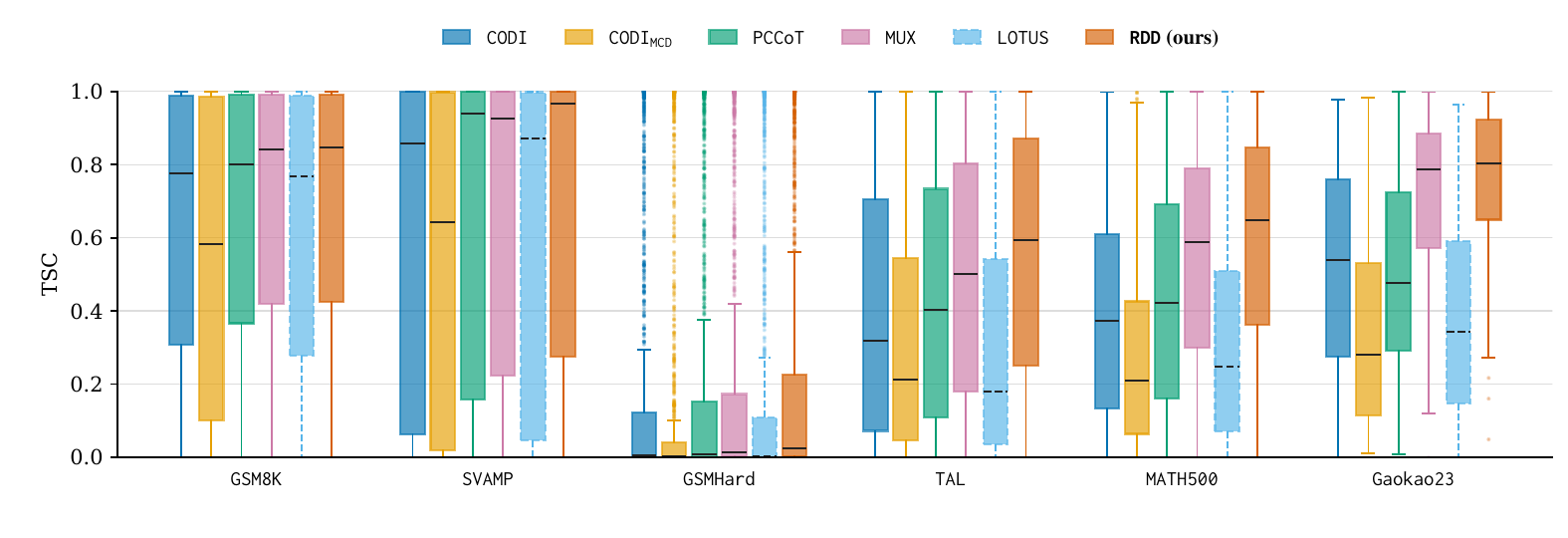}
  \caption{\textbf{Answer-distribution fidelity by benchmark.}
  Per-question JSD (top), TV (middle), and TSC (bottom).
  Lower JSD and TV and higher TSC indicate better fidelity.
  Boxes show the interquartile range with median lines; whiskers extend to observations within $1.5$ times the interquartile range, and dots denote outliers.}
  \label{fig:fidelity-benchmarks}
\end{figure*}

\paragraph{Fidelity analysis by benchmark}
Figure~\ref{fig:fidelity-benchmarks} provides the six-benchmark breakdown of the grouped fidelity analysis in Figure~\ref{fig:fidelity-analysis}.
\texttt{RDD} achieves the lowest mean JSD and TV on all three advanced mathematics benchmarks and the highest mean teacher support coverage on all six benchmarks.

\begin{figure*}[p]
  \centering
  \includegraphics[width=\textwidth]{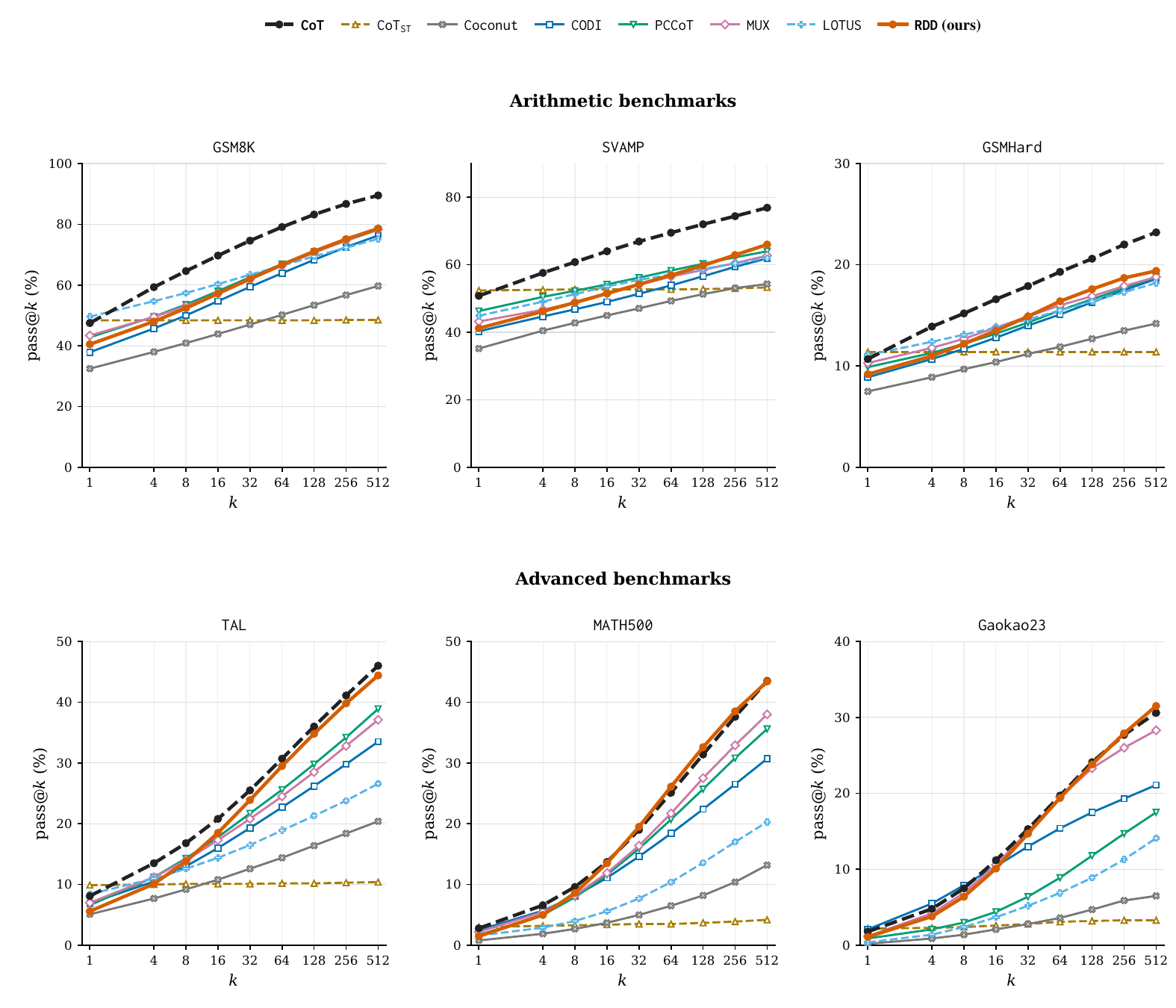}
  \caption{\textbf{Pass@$k$ curves for the methods in Table~\ref{tab:gpt}.}
  Results for \texttt{GPT-2-0.1B} on the arithmetic benchmarks (top row) and advanced benchmarks (bottom row).
  The horizontal axis uses a base-2 logarithmic scale; the vertical axis reports pass@$k$ (\%).
  Markers indicate evaluated $k$ values, with lines connecting adjacent points.}
  \label{fig:passk-main}
\end{figure*}

\begin{figure*}[p]
  \centering
  \includegraphics[width=\textwidth]{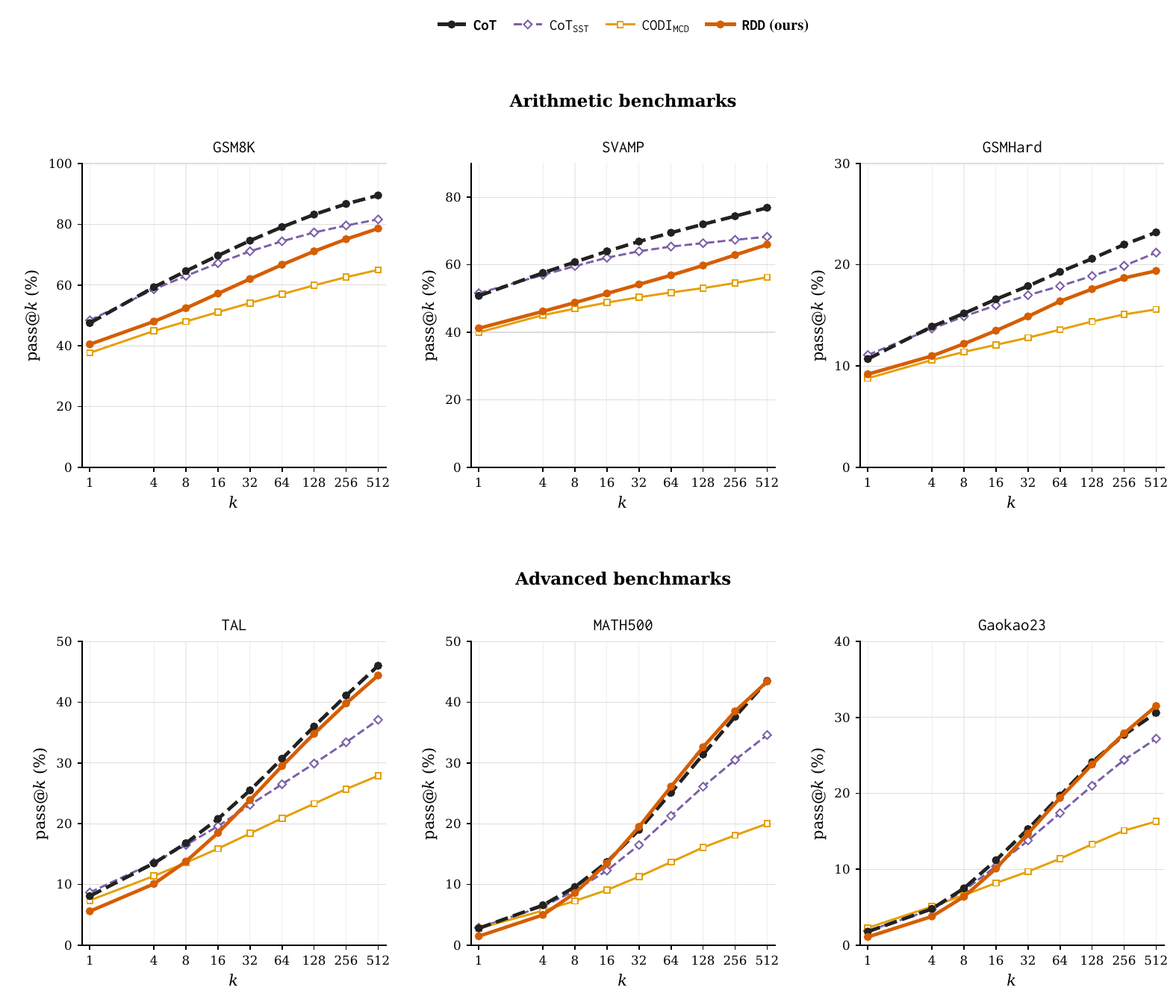}
  \caption{\textbf{Pass@$k$ curves for the methods in Table~\ref{tab:gpt-dist}.}
  Results for \texttt{GPT-2-0.1B} on the arithmetic benchmarks (top row) and advanced benchmarks (bottom row).
  The horizontal axis uses a base-2 logarithmic scale; the vertical axis reports pass@$k$ (\%).
  Markers indicate evaluated $k$ values, with lines connecting adjacent points.}
  \label{fig:passk-distributional}
\end{figure*}

\begin{figure*}[p]
  \centering
  \includegraphics[width=\textwidth]{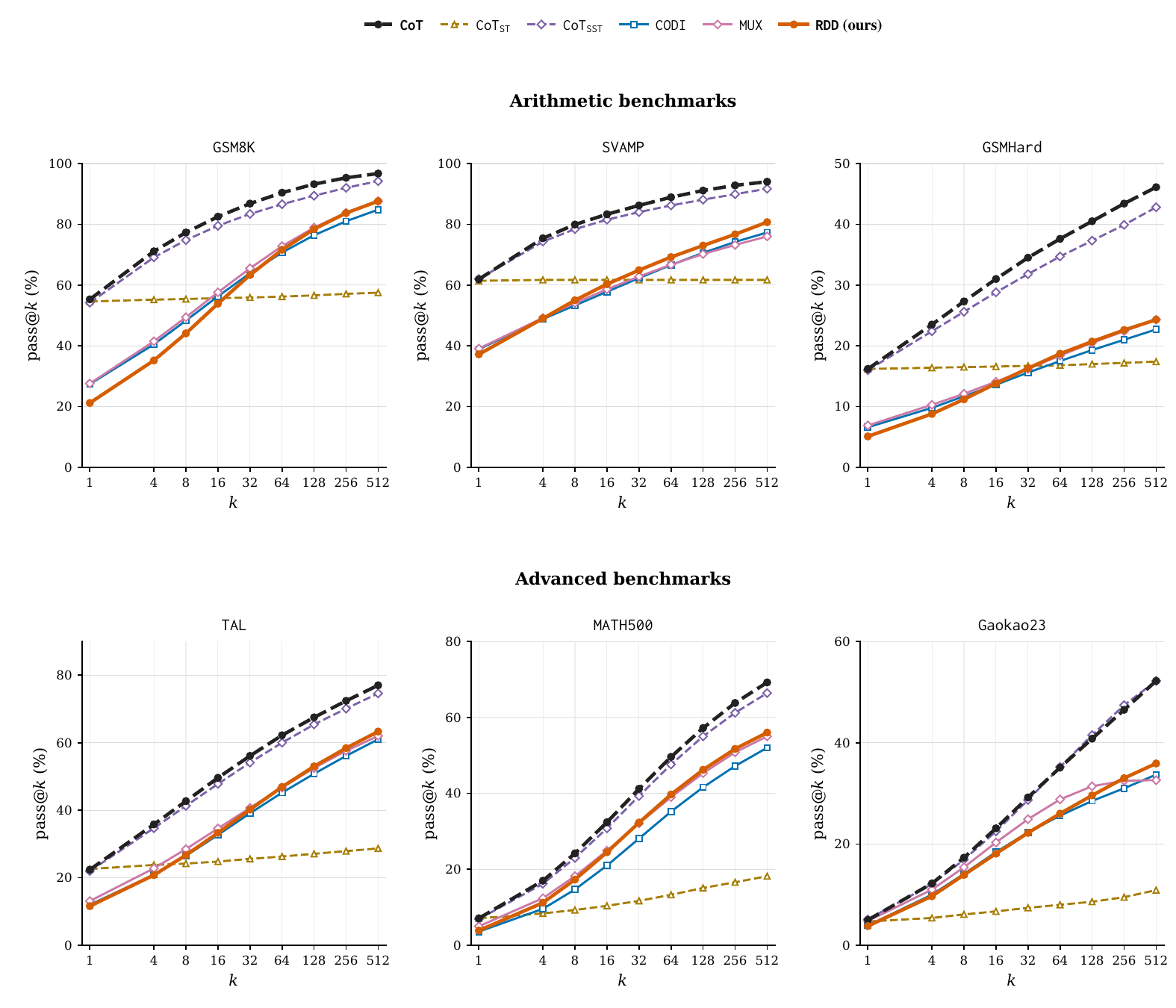}
  \caption{\textbf{Pass@$k$ curves for the methods in Table~\ref{tab:slm}.}
  Results for \texttt{SLM-2-1.7B} on the arithmetic benchmarks (top row) and advanced benchmarks (bottom row).
  The horizontal axis uses a base-2 logarithmic scale; the vertical axis reports pass@$k$ (\%).
  Markers indicate evaluated $k$ values, with lines connecting adjacent points.}
  \label{fig:passk-slm}
\end{figure*}

\paragraph{Answer coverage by benchmark}
Figures~\ref{fig:passk-main}, \ref{fig:passk-distributional}, and~\ref{fig:passk-slm} show the benchmark-wise \texttt{pass@}$k$ curves for the methods in Tables~\ref{tab:gpt}, \ref{tab:gpt-dist}, and~\ref{tab:slm}, respectively, for $k\in\{1,4,8,16,32,64,128,256,512\}$.
\texttt{RDD}'s advantage becomes apparent at larger $k$, particularly on advanced mathematics.
Notably, \texttt{RDD} achieves answer coverage comparable to or even exceeding that of textual \texttt{CoT} on advanced mathematics for \texttt{GPT-2-0.1B}.
Note that \texttt{CoT$_{\texttt{ST}}$}, \texttt{CoT$_{\texttt{SST}}$}, and \texttt{LOTUS} use more latent tokens than other methods.

\section{Declaration of AI Usage}

Large language models were used to assist with manuscript drafting, language editing, proofreading, and the development and debugging of code used in the experiments.
All AI-assisted content and code were reviewed and validated by the authors, who take full responsibility for the accuracy, reliability, and integrity of the final manuscript and experimental results.

\end{document}